\pdfoutput=1
\documentclass[11pt]{article}

\usepackage[final]{acl}
\usepackage{times}
\usepackage{latexsym}

\usepackage[T1]{fontenc}
\usepackage[utf8]{inputenc}

\usepackage{microtype}

\usepackage{amssymb}
\usepackage{multirow}
\usepackage{hyperref}
\usepackage{url}
\usepackage{arydshln}
\usepackage{wrapfig}
\usepackage{amsmath}
\usepackage{colortbl}
\usepackage{xcolor}

\usepackage{enumitem}
\usepackage{tabularx}
\usepackage{adjustbox}
\usepackage{booktabs}

\usepackage{tcolorbox}
\usepackage{pgfplots}
\usepackage{makecell}
\usepackage{color}
\usepackage{circledtext}
\usepackage{textcomp}

\usetikzlibrary{shapes}
\usetikzlibrary{arrows,decorations.pathmorphing,backgrounds,positioning,fit,petri}
\usetikzlibrary{arrows.meta,fit,shapes.arrows}
\usetikzlibrary{positioning,shadows,patterns}
\usepackage{caption}
\usepackage{bm}

\definecolor{tiffanyblue}{RGB}{129,216,208}
\definecolor{bangdiblue}{RGB}{0,149,182}
\definecolor{kleinblue}{RGB}{0,47,167}

\usepackage{tikz}
\usepackage{subfig}
\usepackage{tkz-kiviat,pgfplots}
\pgfplotsset{compat=newest}
\usepgfplotslibrary{patchplots} % 加载 patchplots 库

\usepackage{listings}

\newtcolorbox{promptbox}[2][]{
	width=\linewidth,
	colback = gray!8, 
	colframe = gray!8, 
	boxsep=0pt,left=5pt,right=5pt,top=2pt,bottom=2pt,
	fontupper=\linespread{1.2}\selectfont,
	title=#2,#1,
        fontupper=\small}

\definecolor{myGreen}{RGB}{50, 168, 82}
\definecolor{myYellow}{RGB}{214, 169, 45}
\definecolor{myRed}{RGB}{201, 24, 24}
\definecolor{myBlue}{RGB}{50, 200, 200}
\definecolor{myBrown}{RGB}{140, 50, 50}
\definecolor{myPurple}{RGB}{128, 0, 200} % 定义一种紫色

\definecolor{given}{RGB}{197,217,197}
\definecolor{response}{RGB}{176,224,230}

\title{\textsc{CC-Tuning}: A \underline{\textit{C}}ross-Lingual \underline{\textit{C}}onnection Mechanism for Improving \\ Joint Multilingual Supervised Fine-Tuning}

\author{
  Yangfan Ye$^{1}$,
  Xiaocheng Feng$^{1,2}$\thanks{Corresponding Author},
  Zekun Yuan$^{1}$,
  Xiachong Feng$^{3}$,
  Libo Qin$^{4}$, \\
  \textbf{Lei Huang$^{1}$ 
  Weitao Ma$^{1}$,
  Yichong Huang$^{1}$,
  Zhirui Zhang$^{}$,
  Yunfei Lu$^{5}$}, \\
  \textbf{Xiaohui Yan$^{5}$, 
  Duyu Tang$^{5}$,
  Dandan Tu$^{5}$,
  Bing Qin$^{1,2}$} \\
  $^{1}$Harbin Institute of Technology \quad
  $^{2}$Peng Cheng Laboratory \quad
  $^{3}$The University of Hong Kong \\
  $^{4}$Central South University \quad \quad
  $^{5}$Huawei Technologies Co., Ltd \\
  \texttt{\{yfye,xcfeng,qinb\}@ir.hit.edu.cn}
}

\begin{document}
\maketitle
\begin{abstract}
Current large language models (LLMs) often exhibit imbalanced multilingual capabilities due to their English-centric training corpora.
To address this, existing fine-tuning approaches operating at the \textit{data-level} (e.g., through data augmentation or distillation) typically introduce implicit cross-lingual alignment, overlooking the potential for more profound, \textit{latent-level}\footnote{\textit{latent-level}: referring to direct manipulation of the model's internal representations (e.g., FFN activations)} cross-lingual interactions.
In this work, we propose \textsc{CC-Tuning}, a novel multilingual fine-tuning paradigm that explicitly establishes a cross-lingual connection mechanism at the latent level.
During training, \textsc{CC-Tuning} fuses the feed forward activations from both English and non-English inputs, enabling the model to benefit from both linguistic resources.
This process is facilitated with a trainable \textit{Decision Maker} that identifies beneficial activations.
Furthermore, during inference, a \textit{Transform Matrix} is utilized to simulate the cross-lingual connection under monolingual setting through representation transformation.
Our experiments on six benchmarks covering 22 languages show that \textsc{CC-Tuning} outperforms vanilla SFT and offers a strong latent-level alternative to data-level augmentation methods. Further analysis also highlights the practicality of \textsc{CC-Tuning} and the potential of latent-level cross-lingual interactions in advancing the multilingual performance of LLMs. (Code link: \href{https://github.com/YYF-Tommy/CC-Tuning}{CC-Tuning})
\end{abstract}

\section{Introduction}\label{sec:intro}

Recent advancements in large language models (LLMs) have demonstrated exceptional capabilities in handling diverse tasks~\citep{dong2022survey, wei2022emergent, wei2022chain, shanahan2022talking, zhao2023survey, liu2023pre, huang2025survey} while exhibiting promising generalizability across diverse languages~\citep{ye2023language, qin2024multilingual,huo2025enhancing}. However, significant performance disparities persist across languages due to the overwhelming dominance of English in training corpora, making balanced multilingual proficiency an ongoing research challenge~\citep{touvron2023llama, zhang-etal-2023-dont, ye-etal-2024-globesumm}.

One of the prevailing approaches towards these challenges focuses on joint multilingual supervised fine-tuning (SFT)~\citep{ouyang2022training}, which refers to fine-tuning the model with supervised data spanning multiple languages.
While effective in principle, these methods encounter the ``curse of multilinguality'' – a paradoxical phenomenon where expanding language coverage during joint training leads to performance degradation across both high- and low-resource languages~\citep{conneau-etal-2020-unsupervised, wang-etal-2020-negative}.

\begin{figure}[t]
\includegraphics[width=1\linewidth]{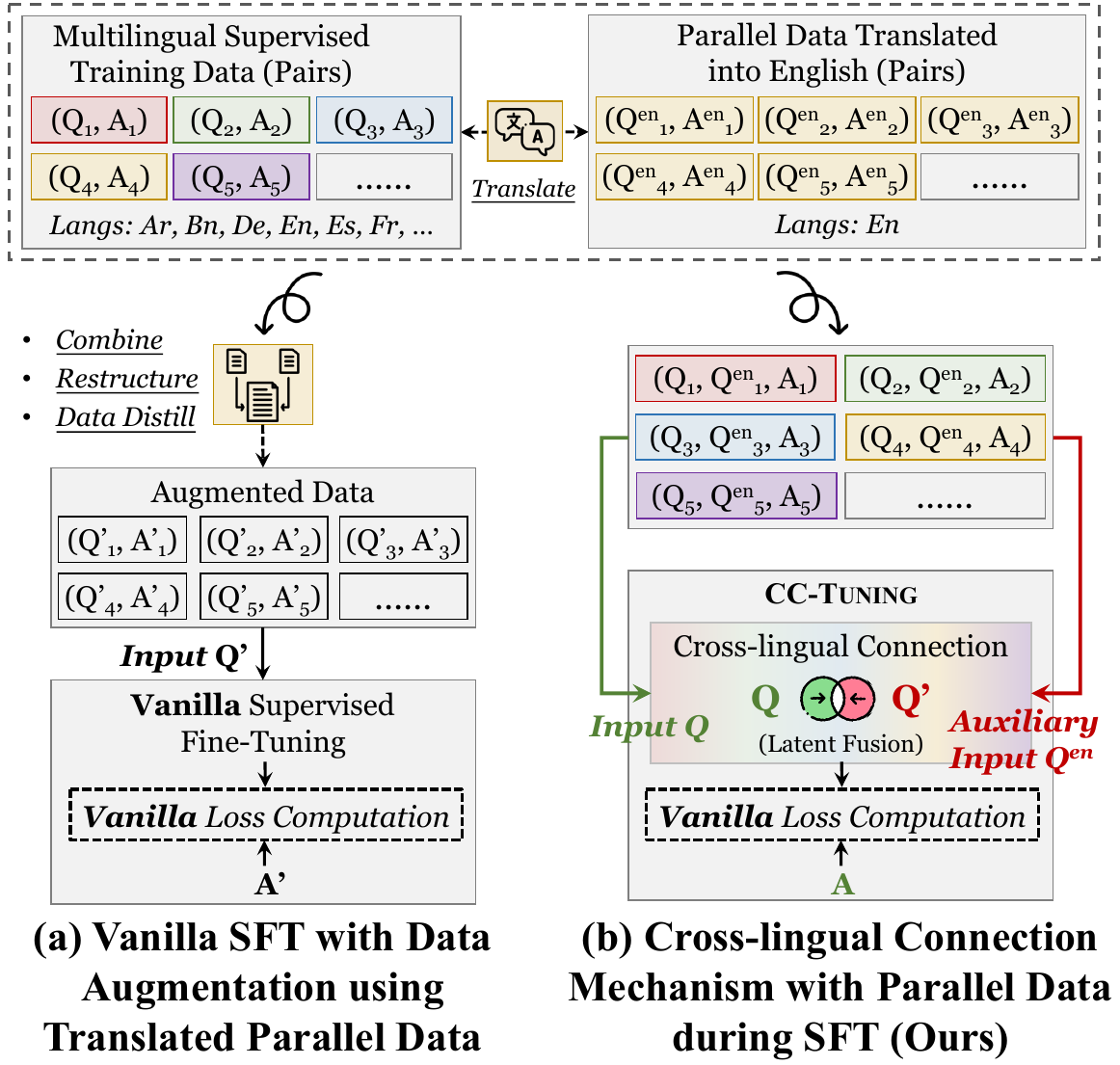}
\caption{Comparison between vanilla supervised fine-tuning with data augmentation at \textbf{data level} (implicit) and our method at \textbf{latent activation level} (explicit).}
\vspace{-1\baselineskip}
\label{fig:diff}
\end{figure}

To address this, current studies primarily focus on data-level interventions through parallel corpus utilization.
Common strategies include: multilingual data augmentation with English-aligned parallel examples~\citep{aharoni-etal-2019-massively, shaham-etal-2024-multilingual}, explicit translation task formulation~\citep{johnson-etal-2017-googles, tang2020multilingual}, and response distillation from resource-rich languages~\citep{zhang2024enhancing}. While these methods demonstrate partial success, their reliance on implicitly introducing data-level text alignment overlooks the potential for deeper, latent-level cross-lingual interactions.

We propose \textsc{CC-Tuning}, a novel multilingual fine-tuning paradigm that introduces \textit{explicit cross-lingual connections} at the latent activation level by fusing feed-forward activations from English and non-English languages (Figure~\ref{fig:diff}). 
This approach is grounded in recent empirical findings highlighting the significant potential of feed-forward activations in improving model's multilingual performance~\citep{ye2024xtransplant}.
During training, our method leverages parallel bilingual inputs and incorporates a trainable \textit{Decision Maker} to identify linguistically beneficial signals from auxiliary English activations, integrating them into the forward propagation of non-English inputs. Additionally, during inference, an ``easy-to-learn'' \textit{Transform Matrix} is utilized to simulate the cross-lingual connection without the parallel bilingual inputs, ensuring the practicality of our approach.
This latent-level interaction mechanism fundamentally differs from conventional data-level approaches, as it establishes direct interlingual activation connections rather than relying on statistical correlations in training data.

To validate our approach, we conduct extensive experiments across six benchmarks encompassing both natural language understanding and generation tasks, spanning 22 languages using two representative LLMs. 
Our results highlight the superiority of \textsc{CC-Tuning} over vanilla SFT in multilingual joint learning scenarios. Besides, compared to data-level augmentation or distillation methods that leverage parallel data, \textsc{CC-Tuning} offers a highly effective alternative for facilitating cross-lingual interaction.
Additionally, our further ablation studies and analysis also provide strong evidence of the practicality and robustness of \textsc{CC-Tuning}.
\section{Related Work}\label{sec:related}

\paragraph{Multilingual Large Language Models.}
Recently, larger models such as Bloom~\citep{scao2022bloom}, Mala-500~\citep{lin2024mala} and Aya Model~\citep{ustun2024aya} have pushed multilingual performance further by leveraging the benefits of greater scale.
Generally, multilingual pretraining and fine-tuning are now the two mainstream methods for improving multilingual capabilities.
Models such as Sabia~\citep{pires2023sabia}, ChineseLLaMA~\citep{cui2023efficient}, ChineseMixtral~\citep{Chinese-Mixtral-8x7B}, PolyLM~\citep{wei2023polylm} and PaLM2~\citep{anil2023palm} have been developed through (continuous) pretraining with large multilingual corpora or language-specific data.
Other models like BLOOMz~\citep{muennighoff2022crosslingual}, m-LLaMA~\citep{zhu2023extrapolating}, Camoscio~\citep{santilli2023camoscio}, Phoenix~\citep{chen2023phoenix} and Bode~\citep{garcia2024introducing} have opted for  leveraging multilingual or language-specific data directly during SFT stage to foster cross-lingual alignment.

\begin{figure*}[t]
    \centering
    \includegraphics[width=1\textwidth]{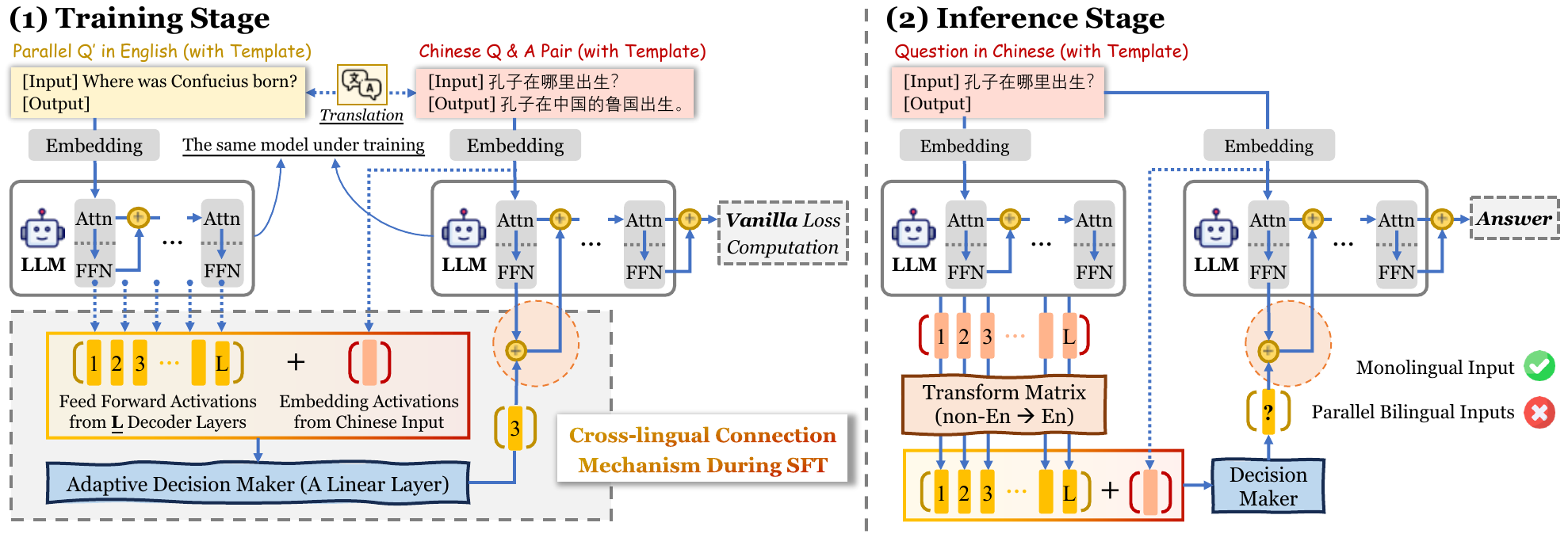}
    \caption{Overview of the cross-lingual connection mechanism in \textsc{CC-Tuning}. In the training stage, \textsc{CC-Tuning} leverages an auxiliary English input alongside the non-English input, while retaining the vanilla loss computation without introducing additional training objectives. In the inference stage, a transform matrix is used to simulate cross-lingual connection in monolingual input scenarios, eliminating the dependence on bilingual parallel input.
    }
    \label{fig:method}
    \vspace{-1\baselineskip}
\end{figure*}

\paragraph{Multilingual Supervised Fine-Tuning.}
Multilingual SFT is an effective way to enhance the multilingual performance of LLMs.
Current research often focuses on data augmentation or distillation techniques to enrich training data and improve model generalization across multiple languages.
For instance, \citet{pan-etal-2024-g} highlighted the importance of diverse, high-quality data for machine translation fine-tuning, while \citet{li2023bactrian} addressed "translationese" by using Google Translate and ChatGPT for multilingual response generation.
In terms of instruction tuning, \citet{shaham-etal-2024-multilingual} showed that adding multilingual examples to English-centric fine-tuning significantly boosts multilingual instruction-following, while \citet{chen-etal-2024-monolingual} demonstrated the superiority of multilingual tuning over language-specific training. Translation-based fine-tuning has been shown to enhance semantic alignment, as argued by \citet{ranaldi-etal-2024-empowering}. Similarly, \citet{zhu2023extrapolating} combined translation data, cross-lingual tasks, and scaling laws to optimize multilingual performance. Additionally, \citet{zhang2024enhancing} proposed a self-distillation approach leveraging LLMs' internal capabilities in resource-rich languages to enhance multilingual performance.

The above methods primarily focus on enriching training data with parallel data to foster implicit cross-lingual alignment. In contrast, our \textsc{CC-Tuning} emphasizes improving the training paradigm by explicitly incorporating cross-lingual latent interactions into the training process.

\section{Method}\label{sec:method}

In this section, we first revisit the vanilla multilingual supervised fine-tuning paradigm, then present the training implementation of \textsc{CC-Tuning} and its specialized configurations during inference stage.

\subsection{Multilingual Supervised Fine-Tuning}
Multilingual supervised fine-tuning enables pretrained models to better perform downstream tasks across diverse languages through training on annotated multilingual instruction dataset $D=\{(x_i,y_i)\}_{i=1}^{N}$, where $N$ represents the size of the dataset, $x_i$ denotes the input question or instruction, and $y_i$ is the corresponding expected output or response. The training process is required to minimize the following objective of negative log-likelihood of the predicted output with respect to the ground-truth response. $\theta$ denotes the parameters of the model.
\begin{equation}
\mathcal{L}_{SFT}(\theta) = \frac{1}{N} \sum_{i=1}^{N}{-\operatorname{log} P(y_i | x_i, \theta)}
\label{eq:loss}
\end{equation}

\paragraph{Data Augmentation with Parallel Data.}
For the multilingual instruction dataset $D$, we define its corresponding English parallel data as $D_{en}$. 
Several previous studies have explored enriching the original training data by merging these two datasets, incorporating additional translation task form data constructed from parallel pairs, or utilizing techniques such as distillation. 
We collectively refer to these augmented datasets as $D_{aug}=\{(x^{aug}_i,y^{aug}_i)\}_{j=1}^{M}$. 
These approaches, in essence, do not alter the SFT process; rather, they introduce additional supervised data, as illustrated below:
\begin{equation}
\begin{aligned}
    &\mathcal{L}_{SFT_{aug}}(\theta) = \frac{1}{N} \sum_{i=1}^{N}{-\operatorname{log} P(y_i | x_i, \theta)} \\
    & \hspace{1.03cm} + \frac{1}{M} \sum_{j=1}^{M}{-\operatorname{log} P(y^{aug}_i | x^{aug}_i, \theta)}
\end{aligned}
\end{equation}

\subsection{\textsc{CC-Tuning}}

We will introduce cross-lingual connection mechanism in \textsc{CC-Tuning} in detail, focusing on its implementation during training and inference stages.

\subsubsection{Training with Cross-lingual Connection}

Motivated by the findings in \citet{ye2024xtransplant}, which empirically demonstrate that feed-forward activations from English hold the potential to significantly enhance a model's performance in non-English languages. The cross-lingual connection mechanism in \textsc{CC-Tuning} aims to incorporate the above latent interactions into the multilingual fine-tuning process, enabling the model to benefit from both English and non-English resources as the parameters are updated.

We denote $D=\{(x_i, y_i)\}_{i=1}^{N}$ as a multilingual supervised instruction dataset, where $x_i$ represents the input question for the $i$-th data point and $y_i$ denotes the corresponding ground-truth response.
Besides, \textsc{CC-Tuning} requires auxiliary parallel data, $D^{en}=\{(x_i, x^{en}_i, y_i)\}_{i=1}^{N}$, where $x^{en}_i$ is the English translation of $x_i$.
Generally, the cross-lingual connection mechanism consists of two key operations: (1) adaptive decision maker and (2) latent feed forward connection.
Notably, these operations are executed just before the \textit{Response Start Token (RST)}, which marks the beginning of the model's response in the training template.
This ensures that our operations can smoothly introduce the intervention into the response generation process.
Assuming the training template is structured as ``\colorbox{gray!20}{\textit{[Input]} \{question\} \textit{[output]} \{answer\}}'', these operations are executed at the position that is right before the \textit{[output]} token.

\paragraph{Adaptive Decision Maker.}
% \paragraph{Beneficial Activation Identification.}
Given an auxiliary input $x^{en}_i$, we first pass it through the model to extract its feed-forward activations $F^{en}_{i}\in \mathbb{R}^{L \times d} = \{f^{en}_{i,l}\}_{l=1}^{L}$ from $L$ decoder layers, where $d$ is the dimensionality of the hidden states. 
Notably, prior research has shown that not all feed-forward activations contribute equally and some may degrade performance~\citep{ye2024xtransplant}.
To mitigate this issue, we introduce a trainable linear layer $W_{DM} \in \mathbb{R}^{d \times L}$, referred to as the \textit{Decision Maker}, which adaptively selects the most beneficial layer. By combining $F^{en}_{i}$ with the embedding activations $e_{i}\in \mathbb{R}^{d}$ of $x_i$, we integrate features from both English and non-English inputs. The resulting combined features are then fed into the \textit{Decision Maker} along with \textit{Gumbel-Softmax}~\citep{jang2016categorical} to achieve the identification as follows:
\begin{equation}
H_i = \frac{1}{L}\sum_{l=1}^{L}{(f^{en}_{i,l} + e_i}) \cdot W_{DM}
\label{eq:dm}
\end{equation}
\begin{equation}
f^{en}_{i,s} = \sum_{l=1}^{L} \operatorname{Gumbel-Softmax}(H_i)_l \cdot f^{en}_{i,l}
\label{eq:gs}
\end{equation}
where $f^{en}_{i,s}\in \mathbb{R}^{d}$ represents the selected activation from the $s$-th layer among the $L$ decoder layers.

\paragraph{Latent Feed Forward Connection.}
The second step aims to transfer the beneficial activation $f^{en}_{i,s}$ identified in the previous step into the forward propagation process of non-English input.
When the input $x_i$ is fed into the model, let the output of all $L$ decoders be denoted as $O_i=\{o_{i,l}\}_{l=1}^{L}$, where each $o_{i,l}$ should have been obtained by combining the feed-forward activations $f_{i,l}$ and self-attention activations $a_{i,l}$ through a residual connection.
However, the incorporation of $f^{en}_{i,s}$ refines this process by connecting itself with the feed-forward activation $f_{i,1}$ from the first decoder layer. Formally, this modification can be expressed as:
\begin{equation}
\tilde{f}_{i,1} = f_{i,1} + f^{en}_{i,s}
\label{eq:fuse}
\end{equation}
The forward propagation of the input $x_i$ then continues with this modification. Consequently, the original decoder outputs $\{o_{i,l}\}_{l=j}^{L}$ will be altered to $\{\tilde{o}_{i,l}\}_{l=j}^{L}$ due to the update of $f_{i,1} \rightarrow \tilde{f}_{i,1}$, leading to new final prediction outcomes $\tilde{o}_{i,L}$.

And within \textsc{CC-Tuning}, the training objective remains the same as the vanilla loss objective in Equation~\ref{eq:loss}. During the tuning process, the model itself, along with the \textit{Decision Maker}, learns to leverage the benefits of both English and non-English resources, improving its multilingual capabilities.

\subsubsection{Inference with Transform Matrix}
Unlike the training stage, our inference process is conducted without the need for parallel inputs. Instead, we leverages a training-free \textit{Transform Matrix} to simulate the cross-lingual connection.

The role of the \textit{Transform Matrix} $W_{T}$ here is to achieve the transformation of $F_{i} = \{f_{i,l}\}_{l=1}^{L} \rightarrow F^{en}_{i} = \{f^{en}_{i,l}\}_{l=1}^{L}$ in the absence of parallel English input $x^{en}_i$.
Specifically, after training, we firstly sample 1,000 parallel pairs $(x_i, x^{en}_i)$ from the datasets $D$ and $D^{en}$, and collect their feed-forward activations, $F_{i}$ and $F^{en}_{i}$, respectively.
These activations are then stacked and denoted as $A=\{f_{i,l}\mid i=1,...,N;l=1,...,L\}$ and $B=\{f^{en}_{i,l}\mid i=1,...,N;l=1,...,L\}$. Therefore, $A$ can be mapped into $B$ as follows through $W_T$:
\begin{equation}
A \cdot W_{T} = B
\end{equation}
To minimize the difference $A$ and $B$, our objective is defined as follows (Least-Squares optimization):
\begin{equation}
    \bm{W}^*_{T}=\mathop{\operatorname{argmin}}_{\bm{W}_{T}} \sum_{i=1}^N\sum_{l=1}^L \left\|f_{i,l}  \bm{W}_{T}-f^{en}_{i,l}\right\|^2
\end{equation}
This problem seeks the optimal $\bm{W}^*_{T}$ that minimizes the distance between the source and target representations. Hence, the closed-form solution to this optimization problem is:
\begin{equation}
    \small
    \bm{W}^*_{T} = \left(\sum_{i=1}^{N}\sum_{i=l}^{L} (f_{i,l})^T f_{i,l} \right)^{-1} \left(\sum_{i=1}^{N}\sum_{i=l}^{L} (f_{i,l})^T f^{en}_{i,l}\right)
\end{equation}
Once the optimal $\bm{W}_{T}$ has been learned, it can be applied to the non-English representation to map it to the corresponding English representation.
This resulting mapped representation $F_{i} \cdot W_{T}$, then substitutes $F^{en}_{i}=\{f^{en}_{i,l}\}_{l=1}^{L}$ in equations~\ref{eq:dm}, \ref{eq:gs}, \ref{eq:fuse}, thereby simulating the cross-lingual connection.
This alignment effectively eliminates the dependence for bilingual parallel data and enables the simulation of cross-lingual connection in a monolingual scenario.

\section{Experiments}\label{sec:exp}

\definecolor{up-green}{RGB}{0,120,0}
\definecolor{down-red}{RGB}{255,0,0}

\begin{table*}[t]
\small
\renewcommand{\arraystretch}{1.2} % 设置行间距为默认的 1.5 倍
\setlength{\dashlinedash}{4pt} % 设置虚线段长度
\setlength{\dashlinegap}{2pt}  % 设置虚线段之间的间隔
  \centering
  \resizebox{\linewidth}{!}{
    \begin{tabular}{lcccccccccccc}
    \toprule
    \multirow{3}[4]{*}{\textbf{Method}} & \multicolumn{6}{c}{\textbf{Multilingual Understanding}} & \multicolumn{6}{c}{\textbf{Multilingual Generation}} \\
    \cmidrule(lr){2-7}
    \cmidrule(lr){8-13}
    & \multicolumn{2}{c}{\textbf{XNLI}} & \multicolumn{2}{c}{\textbf{XStoryCloze}} & \multicolumn{2}{c}{\textbf{MMMLU}} & \multicolumn{2}{c}{\textbf{MKQA}} & \multicolumn{2}{c}{\textbf{XQuAD}} & \multicolumn{2}{c}{\textbf{XLSum}}\\
    \cmidrule(lr){2-3}
    \cmidrule(lr){4-5}
    \cmidrule(lr){6-7}
    \cmidrule(lr){8-9}
    \cmidrule(lr){10-11}
    \cmidrule(lr){12-13}
    & \multicolumn{1}{c}{LLaMA.} & \multicolumn{1}{c}{Qwen.} & \multicolumn{1}{c}{LLaMA.} & \multicolumn{1}{c}{Qwen.} & \multicolumn{1}{c}{LLaMA.} & \multicolumn{1}{c}{Qwen.} & \multicolumn{1}{c}{LLaMA.} & \multicolumn{1}{c}{Qwen.} & \multicolumn{1}{c}{LLaMA.} & \multicolumn{1}{c}{Qwen.} & \multicolumn{1}{c}{LLaMA.} & \multicolumn{1}{c}{Qwen.} \\
    \midrule
    % \headercolor
    \multicolumn{13}{c}{\textbf{\textit{Baselines}}} \\
    \midrule
    % \underline{64.1} & 28.7
    % Vanilla Model & 25.62 & 45.83 & 35.54 & 67.09 & 20.76 & 46.53 & 7.56 & 9.05 & 55.67 & 64.11 & 5.71 & 5.08 \\
    % \midrule
    \textsc{Ml-SFT} & 31.88 & 48.23 & 65.23 & 70.06 & 40.20 & \textbf{50.05} & 14.64 & 14.73 & 60.42 & 63.61 & 12.27 & 12.40 \\
    \cdashline{1-13}\noalign{\vskip 0.4ex}
    \textsc{\quad+En} & 35.02 & 50.76 & 65.13 & 71.63 & 39.62 & 48.80 & 13.28 & 13.05 & 57.40 & 62.34 & 12.04 & 12.20 \\
    \textsc{\quad+MT} & 35.90 & 47.05 & 69.90 & 70.50 & 40.68 & 47.49 & 13.56 & 13.54 & 58.40 & 64.03 & 12.89 & 12.48 \\
    \textsc{\quad+SDRRL} & 29.74 & 52.36 & 55.82 & 80.67 & 28.06 & 47.28 & -- & -- & -- & -- & -- & -- \\
    \midrule
    % \headercolor
    \multicolumn{13}{c}{\textbf{\textit{Ours}}} \\
    \midrule
    \textsc{CC-Tuning} & \cellcolor{cyan!15}\textbf{38.42} & \cellcolor{cyan!15}51.00 & \cellcolor{cyan!15}70.60 & \cellcolor{cyan!15}71.43 & \cellcolor{cyan!15}\textbf{40.74} & \cellcolor{gray!15}49.65 & \cellcolor{cyan!15}\textbf{15.94} & \cellcolor{cyan!15}\textbf{14.84} & \cellcolor{cyan!15}\textbf{61.85} & \cellcolor{cyan!15}63.72 & \cellcolor{cyan!15}12.88 & \cellcolor{cyan!15}12.50 \\
    & \cellcolor{cyan!15}{(+6.54)} & \cellcolor{cyan!15}{(+2.77)} & \cellcolor{cyan!15}{(+5.37)} & \cellcolor{cyan!15}{(+1.37)} & \cellcolor{cyan!15}{(+0.54)} & \cellcolor{gray!15}{(-0.40)} & \cellcolor{cyan!15}{(+1.30)} & \cellcolor{cyan!15}{(+0.11)} & \cellcolor{cyan!15}{(+1.21)} & \cellcolor{cyan!15}{(+0.11)} & \cellcolor{cyan!15}{(+0.61)} & \cellcolor{cyan!15}{(+0.10)} \\
    \cdashline{1-13}\noalign{\vskip 0.4ex}
    \textsc{\quad+En} & \cellcolor{gray!15}32.72 & \cellcolor{gray!15}49.48 & \cellcolor{gray!15}60.94 & \cellcolor{gray!15}64.69 & \cellcolor{gray!15}38.73 & \cellcolor{gray!15}47.35 & \cellcolor{cyan!15}14.61 & \cellcolor{cyan!15}13.56 & \cellcolor{cyan!15}60.89 & \cellcolor{cyan!15}62.69 & \cellcolor{cyan!15}12.78 & \cellcolor{cyan!15}12.63 \\
    & \cellcolor{gray!15}{(-2.30)} & \cellcolor{gray!15}{(-1.28)} & \cellcolor{gray!15}{(-4.19)} & \cellcolor{gray!15}{(-6.94)} & \cellcolor{gray!15}{(-0.89)} & \cellcolor{gray!15}{(-1.45)} & \cellcolor{cyan!15}{(+1.33)} & \cellcolor{cyan!15}{(+0.51)} & \cellcolor{cyan!15}{(+3.40)} & \cellcolor{cyan!15}{(+0.35)} & \cellcolor{cyan!15}{(+0.74)} & \cellcolor{cyan!15}{(+0.43)} \\
    \textsc{\quad+MT} & \cellcolor{cyan!15}36.44 & \cellcolor{cyan!15}48.13 & \cellcolor{cyan!15}\textbf{73.54} & \cellcolor{cyan!15}71.39 & \cellcolor{gray!15}38.87 & \cellcolor{cyan!15}49.39 & \cellcolor{cyan!15}15.59 & \cellcolor{cyan!15}13.77 & \cellcolor{cyan!15}61.55 & \cellcolor{cyan!15}\textbf{64.26} & \cellcolor{cyan!15}\textbf{13.05} & \cellcolor{cyan!15}\textbf{12.87} \\
    & \cellcolor{cyan!15}{(+0.54)} & \cellcolor{cyan!15}{(+1.08)} & \cellcolor{cyan!15}{(+3.64)} & \cellcolor{cyan!15}{(+0.89)} & \cellcolor{gray!15}{(-1.81)} & \cellcolor{cyan!15}{(+1.90)} & \cellcolor{cyan!15}{(+2.03)} & \cellcolor{cyan!15}{(+0.23)} & \cellcolor{cyan!15}{(+3.10)} & \cellcolor{cyan!15}{(+0.23)} & \cellcolor{cyan!15}{(+0.16)} & \cellcolor{cyan!15}{(+0.39)} \\
    \textsc{\quad+SDRRL} & \cellcolor{cyan!15}29.84 & \cellcolor{cyan!15}\textbf{53.06} & \cellcolor{cyan!15}69.19 & \cellcolor{cyan!15}\textbf{80.93} & \cellcolor{cyan!15}37.77 & \cellcolor{cyan!15}47.87 & -- & -- & -- & -- & -- & -- \\
    & \cellcolor{cyan!15}{(+0.10)} & \cellcolor{cyan!15}{(+0.70)} & \cellcolor{cyan!15}{(+13.37)} & \cellcolor{cyan!15}{(+0.26)} & \cellcolor{cyan!15}{(+9.71)} & \cellcolor{cyan!15}{(+0.59)} & -- & -- & -- & -- & -- & -- \\
    \bottomrule
    \end{tabular}
  }
  \caption{Main results that are the averages of the performance across all languages involved for each dataset. \colorbox{cyan!15}{Blue cell} indicates better performance than the vanilla \textsc{Ml-SFT} under the same training data setting, while \colorbox{gray!15}{Gray cell} indicates the opposite. \textbf{Bold} numbers indicate the best performance. LLaMA. and Qwen. respectively represent \textit{LLaMA-3.1-8B} and \textit{Qwen2.5-7B}.}
  \label{tab:main}
\end{table*}

\subsection{Setup}

\paragraph{Models.}
We selected two representative LLMs: (1) \textit{LLaMA-3.1-8B}~\citep{dubey2024llama} and (2) \textit{Qwen2.5-7B}~\citep{qwen2.5}.

\paragraph{Training Corpus.}
We totally select 20,236 multilingual instruction pairs from \textit{aya dataset}~\citep{singh2024aya} as our training corpus and the multilingual training corpus covers more than 60 languages, ensuring extensive multilingual coverage. Our training processes are conducted on \textit{8 * A800-SXM4-80GB} with the following settings: \textit{batch size=16}, \textit{epochs=3}, \textit{learning rate=1.0e-5}, \textit{warmup ratio=0.1}, and \textit{bf16=true}. The implementation is based on \textit{LLaMA-Factory}~\citep{zheng2024llamafactory}.

\paragraph{Baselines.}
More details are in Appendix~\ref{app:baselines}.
\begin{itemize}[leftmargin=*]
\setlength{\parsep}{0pt}
\setlength{\parskip}{0pt}
\item \textbf{\textsc{Ml-SFT}} represents vanilla supervised instruction tuning~\citep{ouyang2022training} with original multilingual instruction dataset (data size=$N$).
\item \textbf{\textsc{Ml-SFT+En}} incorporates the full parallel English version of the dataset for training, followed by vanilla supervised fine-tuning (data size=$2N$). 

\item \textbf{\textsc{Ml-SFT+MT}} constructs additional translation task form data by pairing the original multilingual instruction dataset with its parallel English version and then applies supervised instruction tuning (data size=$2N$).

\item \textbf{\textsc{Ml-SFT+SDRRL}}~\citep{zhang2024enhancing} is a self-distillation-based method that integrates English instruction tuning data and its multilingual code-switched extensions. Additionally, it incorporates partially translated data and completion data for fine-tuning (LLaMA-3.1-8B: data size$\approx$$1.2N$, Qwen2.5-7B: data size$\approx$$1.6N$).
\end{itemize}
And \textbf{\textsc{CC-Tuning (+En, +MT, +SDRRL)}} refers to our method applying the cross-lingual connection mechanism and its combination with different above mentioned training data settings.

\paragraph{Evaluation Datasets.}
We conduct experiments on 6 benchmarks, which can be categorized into:
\begin{itemize}[leftmargin=*]
\setlength{\parsep}{0pt}
\setlength{\parskip}{0pt}
\item \textbf{Multilingual Understanding:} (1) \textit{XNLI}~\citep{conneau2018xnli}, a multilingual natural language inference (NLI) dataset, (2) \textit{XStoryCloze}~\citep{lin-etal-2022-shot}, a multilingual commonsense reasoning dataset for evaluating story understanding and (3) \textit{MMMLU}, the multilingual version of \textit{MMLU}~\citep{hendrycks2020measuring}, designed to evaluate models' general knowledge.
\item \textbf{Multilingual Generation:} (1) \textit{MKQA}~\citep{longpre-etal-2021-mkqa}, an open-domain multilingual question answering evaluation dataset, (2) \textit{XQuAD}~\citep{artetxe-etal-2020-cross}, a question answering dataset and (3) \textit{XLSum}~\citep{hasan-etal-2021-xl}, a multilingual abstractive summarization benchmark comprising professionally annotated article-summary pairs.
\end{itemize}
For each of the above datasets, we conduct experiments on 10 language subsets, covering a total of 22 languages.
For \textit{XNLI}, \textit{XStoryCloze}, \textit{MMMLU}, \textit{MKQA} and \textit{XQuAD} datasets, \textit{Accuracy} metric is used for evaluation. And for \textit{XLSum} dataset, \textit{ROUGE-L} scores are reported.
We use greedy decoding with a max of 40 new tokens for each model.
Detailed information on the datasets and evaluations can be found in Appendix~\ref{app:dataNeval}.

\subsection{Main Results}
The average results across the different languages involved in each dataset are presented in Table~\ref{tab:main}.
The detailed results for different languages can be found in Table~\ref{tab:main_lang_nlu}, \ref{tab:main_lang_nlg}.
Note that the results of applying \textbf{\textsc{+SDRRL}} to NLG tasks are not reported, as it may lead to deviations from the prompt language in model responses, as shown in Appendix~\ref{app:NR}.

\paragraph{(1) \textsc{CC-Tuning} outperforms vanilla SFT in joint multilingual learning scenarios.}
The results in Table~\ref{tab:main} demonstrate that under the same multilingual training data settings of original data, \textsc{+MT} and \textsc{+SDRRL}, \textsc{CC-Tuning} significantly outperforms vanilla SFT in both multilingual understanding and multilingual generation tasks. 
However, under the \textsc{+En} setting, where more than half of the training data is in English, the cross-lingual connection becomes an \textsc{En2En} connection. 
This shift undermines the core goal of \textsc{CC-Tuning}—to promote cross-lingual latent interaction—leading to a notable decline in performance, which also emphasizes \textsc{CC-Tuning}'s alignment with its motivation and use case in joint multilingual learning scenarios.

\paragraph{(2) \textsc{CC-Tuning} with original training data outperforms data augmentation and distillation methods on \textit{LLaMA-3.1-8B}.}
As observed on \textit{LLaMA-3.1-8B}, \textsc{CC-Tuning}, even when trained solely with the original dataset (data size = $N$), outperforms the data augmentation and distillation approaches of \textsc{Ml-SFT+En} (data size = $2N$), \textsc{+MT} (data size = $2N$), and \textsc{+SDRRL} (data size $\approx$ $1.2N$), which utilize larger training set.
This suggests that, compared to implicitly introducing cross-lingual alignment information at the data level, the explicit latent-level cross-lingual connection mechanism in \textsc{CC-Tuning} provides a compelling alternative for facilitating cross-lingual interaction.

% Ablation Studies
\subsection{Ablation Studies}
We perform ablation studies to assess the following aspects: (1) the effectiveness of the \textit{Transform Matrix}, (2) the necessity of the \textit{Decision Maker}, and (3) the advantages of feed-forward activations in facilitating cross-lingual interactions.

% MSE：Mean Squared Error
\begin{table}[t]
\small
\renewcommand{\arraystretch}{1.2} % 设置行间距为默认的 1.5 倍
  \centering
  \resizebox{\columnwidth}{!}{
    \begin{tabular}{lcccccc}
    \toprule
    \multirow{2}{*}{\parbox{1.9cm}{\quad Method\\($|M|=1000$)}} & {XNLI} & {XStoryCloze} & {MMMLU} & {MKQA} & {XQuAD} & {XLSum} \\
    \cmidrule(lr){2-7}
     & \multicolumn{6}{c}{\textit{Model: LLaMA-3.1-8B}} \\
    \midrule
    \quad \textit{MSE} value & \multicolumn{6}{c}{\textit{MSE = $\frac{1}{N\times L}\sum_{i=1}^{N}\sum_{l=1}^{L}(\frac{1}{d}\|f_{i,l}\cdot W_{T} - f^{en}_{i,l}\|_2^2)$}} \\
    \midrule
    % \noalign{\vskip 0.6ex}
    \textsc{CC-Tuning} & 0.012427 & 0.013256 & 0.013196 & 0.021572 & 0.015428 & 0.014314 \\
    \textsc{\quad+En} & 0.014215 & 0.012734 & 0.012917 & 0.018137 & 0.015919 & 0.014046 \\
    \textsc{\quad+MT} & 0.020413 & 0.021251 & 0.023016 & 0.027770 & 0.021769 & 0.025639 \\
    \textsc{\quad+SDRRL} & 0.017896 & 0.019859 & 0.017098 & -- & -- & -- \\
    \midrule
    \quad\textsc{Avg.MSE} & 0.016238 & 0.016775 & 0.016557 & 0.022493 & 0.017705 & 0.018000 \\
    \bottomrule
    \toprule
    \quad $|\Delta|$ value & \multicolumn{6}{c}{\textit{$|\Delta|$ = $|$ Result(Parallel Bilingual Input) - Result(Transform Matrix) $|$}} \\
    \midrule
    % \noalign{\vskip 0.6ex}
    \textsc{CC-Tuning} & 0.16 & 0.60 & 0.03 & 0.01 & 0.21 & 0.28 \\
    \textsc{\quad+En} & 0.08 & 0.31 & 0.25 & 0.01 & 0.08 & 0.16 \\
    \textsc{\quad+MT} & 0.01 & 0.23 & 0.36 & 0.12 & 0.07 & 0.10 \\
    \textsc{\quad+SDRRL} & 0.06 & 0.56 & 0.11 & -- & -- & -- \\
    \midrule
    \quad\textsc{Avg.}$|\Delta|$ & 0.08 & 0.43 & 0.19 & 0.05 & 0.12 & 0.18 \\
    \bottomrule
    \end{tabular}
  }
  \caption{The results of mean squared error between feed-forward representations in English and the transformed representations after applying the \textit{Transform Matrix}, as well as the performance difference $|\Delta|$ between using parallel bilingual inputs and applying \textit{Transform Matrix}.}
  \label{tab:align}
\end{table}

\paragraph{(1) The \textit{Transform Matrix} aligns well with the effect of using parallel bilingual inputs.}
We verify whether the \textit{Transform Matrix} $W_{T}$ can effectively achieve the alignment by evaluating the mean squared error (MSE) between $f_{i,l}\cdot W_{T}$ and $f^{en}_{i,l}$ as well as the performance difference $|\Delta|$ between using parallel bilingual inputs during inference and applying the \textit{Transform Matrix}.
The results in Table~\ref{tab:align} show that the MSE value reaches the order of magnitude as low as $10^{-2}$, indicating that the \textit{Transform Matrix} effectively transforms $f_{i,l}$ into $f^{en}_{i,l}$. Additionally, the small performance difference $|\Delta|$ further suggests that the \textit{Transform Matrix} serves as an effective substitute for parallel bilingual inputs, achieving great alignment.

\definecolor{my-blue}{RGB}{100,149,237}  % 类似的柔和蓝色
\definecolor{my-purple}{RGB}{221,160,221}  % 类似的柔和紫色
\definecolor{my-gray}{RGB}{192,192,192}   % 类似的柔和橙色

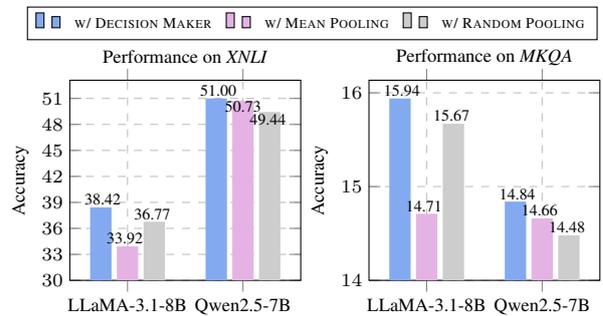
\begin{figure}[t]
\centering
\pgfplotsset{width=0.6\linewidth, height=0.55\linewidth, compat=1.15}
\footnotesize
\begin{tikzpicture}
\scriptsize{
\begin{axis}[
    title={Performance on \textit{XNLI}},
    title style={yshift=-0.3em},
    at={(0em, 0em)},
    ymajorgrids=true,
    xmajorgrids=true,
    grid style=dashed,
    ybar, % 设置柱子宽度
    bar width=8pt, % 设置柱子的宽度
    symbolic x coords={LLaMA-3.1-8B, Qwen2.5-7B}, % X 轴的分组
    xtick=data, % 根据数据自动生成 X 轴刻度
    ymin=30, % Y 轴最小值
    ylabel={Accuracy}, % Y 轴标签
    ytick distance=3, % 设置 Y 轴刻度步长为 2
    ytick align=inside,
    enlarge x limits=0.5, % 增加 X 轴范围，避免柱子被挡住
    legend style={
        at={(1.05,1.2)},
        anchor=south,
        legend columns=3,
        nodes={scale=0.8, transform shape},
        column sep=0.5em, % 设置列间隔为1em（可以根据需要调整）
        },
    ]
    \addplot[fill=my-blue!80, draw=none] coordinates {(LLaMA-3.1-8B, 38.42) (Qwen2.5-7B, 51.00)};
    \addlegendentry{\textsc{w/ Decision Maker}}
    % 手动添加数值标签
    \node[font=\tiny, yshift=3pt, anchor=east] at (axis cs:LLaMA-3.1-8B, 38.42) {38.42};
    \node[font=\tiny, yshift=3pt, anchor=east] at (axis cs:Qwen2.5-7B, 51.00) {51.00};

    % 第二个柱子
    \addplot[fill=my-purple!80, draw=none] coordinates {(LLaMA-3.1-8B, 33.92) (Qwen2.5-7B, 50.73)};
    \addlegendentry{\textsc{w/ Mean Pooling}}
    % 手动添加数值标签
    \node[font=\tiny, yshift=3pt, xshift=-1pt] at (axis cs:LLaMA-3.1-8B, 33.92) {33.92};
    \node[font=\tiny, yshift=-2pt] at (axis cs:Qwen2.5-7B, 50.73) {50.73};
    
    \addplot[fill=my-gray!80, draw=none] coordinates {(LLaMA-3.1-8B, 36.77) (Qwen2.5-7B, 49.44)};
    \addlegendentry{\textsc{w/ Random Pooling}}
    % 手动添加数值标签
    \node[font=\tiny, yshift=3pt, anchor=west] at (axis cs:LLaMA-3.1-8B, 36.77) {36.77};
    \node[font=\tiny, yshift=-3pt, anchor=west] at (axis cs:Qwen2.5-7B, 49.44) {49.44};
    
    \end{axis}

\begin{axis}[
    title={Performance on \textit{MKQA}},
    title style={yshift=-0.4em},
    at={(16em,0em)},
    ymajorgrids=true,
    xmajorgrids=true,
    grid style=dashed,
    ybar, % 设置柱子宽度
    bar width=8pt, % 设置柱子的宽度
    symbolic x coords={LLaMA-3.1-8B, Qwen2.5-7B}, % X 轴的分组
    xtick=data, % 根据数据自动生成 X 轴刻度
    ymin=14, % Y 轴最小值
    ylabel={Accuracy}, % Y 轴标签
    ytick distance=1, % 设置 Y 轴刻度步长为 2
    ytick align=inside,
    enlarge x limits=0.5, % 增加 X 轴范围，避免柱子被挡住
    ]
    \addplot[fill=my-blue!80, draw=none] coordinates {(LLaMA-3.1-8B, 15.94) (Qwen2.5-7B, 14.84)};
    % 手动添加数值标签
    \node[font=\tiny, yshift=3pt, anchor=east] at (axis cs:LLaMA-3.1-8B, 15.94) {15.94};
    \node[font=\tiny, yshift=3pt, anchor=east] at (axis cs:Qwen2.5-7B, 14.84) {14.84};
    
    \addplot[fill=my-purple!80, draw=none] coordinates {(LLaMA-3.1-8B, 14.71) (Qwen2.5-7B, 14.66)};
    % 手动添加数值标签
    \node[font=\tiny, yshift=3pt, xshift=-1pt] at (axis cs:LLaMA-3.1-8B, 14.71) {14.71};
    \node[font=\tiny, yshift=3pt, xshift=-1pt] at (axis cs:Qwen2.5-7B, 14.66) {14.66};
    
    \addplot[fill=my-gray!80, draw=none] coordinates {(LLaMA-3.1-8B, 15.67) (Qwen2.5-7B, 14.48)};
    % 手动添加数值标签
    \node[font=\tiny, yshift=3pt, anchor=west] at (axis cs:LLaMA-3.1-8B, 15.67) {15.67};
    \node[font=\tiny, yshift=3pt, anchor=west] at (axis cs:Qwen2.5-7B, 14.48) {14.48};
    
    \end{axis}
    }
    \end{tikzpicture}
    \caption{Performance comparisons of using \textit{Decision Maker}, \textit{Mean Pooling} and \textit{Random Pooling} strategy on \textit{XNLI} and \textit{MKQA} datasets.
    }
    \label{fig:dm_adv}
\end{figure}

\paragraph{(2) The \textit{Decision Maker} plays a crucial role.}
To verify the necessity of the \textit{Decision Maker}, we replaced it with two alternative strategies—\textit{Mean Pooling} and \textit{Random Pooling}—during both training and inference, and compared their performance in Figure~\ref{fig:dm_adv}.
In \textit{Mean Pooling}, the feed-forward activations from all layers are averaged, while in \textit{Random Pooling}, a single activation is randomly selected from the set of feed-forward activations across all layers. 
The results demonstrate that the performance with the \textit{Decision Maker} significantly outperforms the other two strategies, confirming that the \textit{Decision Maker} effectively serves its role in beneficial activation identification and contributes to the overall training paradigm of \textsc{CC-Tuning}.

\definecolor{my-green}{RGB}{72,209,204}
\definecolor{my-red}{RGB}{255,192,203}
\definecolor{my-yellow}{RGB}{255,230,180}

\begin{figure}[t]
\centering
\pgfplotsset{width=0.6\linewidth, height=0.55\linewidth, compat=1.15}
\footnotesize
\begin{tikzpicture}
\scriptsize{
\begin{axis}[
    title={Performance on \textit{XNLI}},
    title style={yshift=-0.3em},
    at={(0em, 0em)},
    ymajorgrids=true,
    xmajorgrids=true,
    grid style=dashed,
    ybar, % 设置柱子宽度
    bar width=8pt, % 设置柱子的宽度
    symbolic x coords={LLaMA-3.1-8B, Qwen2.5-7B}, % X 轴的分组
    xtick=data, % 根据数据自动生成 X 轴刻度
    ymin=30, % Y 轴最小值
    ylabel={Accuracy}, % Y 轴标签
    ytick distance=3, % 设置 Y 轴刻度步长为 2
    ytick align=inside,
    enlarge x limits=0.5, % 增加 X 轴范围，避免柱子被挡住
    legend style={
        at={(1.05,1.2)},
        anchor=south,
        legend columns=3,
        nodes={scale=0.8, transform shape},
        column sep=0.5em, % 设置列间隔为1em（可以根据需要调整）
        },
    ]
    \addplot[fill=my-green, draw=none] coordinates {(LLaMA-3.1-8B, 38.42) (Qwen2.5-7B, 51.00)};
    \addlegendentry{\textsc{CC-Tuning (FFN)}}
    % 手动添加数值标签
    \node[font=\tiny, yshift=3pt, anchor=east] at (axis cs:LLaMA-3.1-8B, 38.42) {38.42};
    \node[font=\tiny, yshift=3pt, anchor=east] at (axis cs:Qwen2.5-7B, 51.00) {51.00};

    % 第二个柱子
    \addplot[fill=my-red, draw=none] coordinates {(LLaMA-3.1-8B, 36.31) (Qwen2.5-7B, 50.34)};
    \addlegendentry{\textsc{CC-Tuning (Attn)}}
    % 手动添加数值标签
    \node[font=\tiny, yshift=3pt] at (axis cs:LLaMA-3.1-8B, 36.31) {36.31};
    \node[font=\tiny, yshift=-1pt] at (axis cs:Qwen2.5-7B, 50.34) {50.34};
    
    \addplot[fill=my-yellow, draw=none] coordinates {(LLaMA-3.1-8B, 34.37) (Qwen2.5-7B, 50.60)};
    \addlegendentry{\textsc{CC-Tuning (Block)}}
    % 手动添加数值标签
    \node[font=\tiny, yshift=3pt, anchor=west] at (axis cs:LLaMA-3.1-8B, 34.37) {34.37};
    \node[font=\tiny, yshift=-7pt, anchor=west] at (axis cs:Qwen2.5-7B, 50.60) {50.60};
    
    \end{axis}

\begin{axis}[
    title={Performance on \textit{MKQA}},
    title style={yshift=-0.4em},
    at={(16em,0em)},
    ymajorgrids=true,
    xmajorgrids=true,
    grid style=dashed,
    ybar, % 设置柱子宽度
    bar width=8pt, % 设置柱子的宽度
    symbolic x coords={LLaMA-3.1-8B, Qwen2.5-7B}, % X 轴的分组
    xtick=data, % 根据数据自动生成 X 轴刻度
    ymin=14, % Y 轴最小值
    ylabel={Accuracy}, % Y 轴标签
    ytick distance=1, % 设置 Y 轴刻度步长为 2
    ytick align=inside,
    enlarge x limits=0.5, % 增加 X 轴范围，避免柱子被挡住
    ]
    \addplot[fill=my-green, draw=none] coordinates {(LLaMA-3.1-8B, 15.94) (Qwen2.5-7B, 14.84)};
    % 手动添加数值标签
    \node[font=\tiny, yshift=3pt, anchor=east] at (axis cs:LLaMA-3.1-8B, 15.94) {15.94};
    \node[font=\tiny, yshift=3pt, anchor=east] at (axis cs:Qwen2.5-7B, 14.84) {14.84};
    
    \addplot[fill=my-red, draw=none] coordinates {(LLaMA-3.1-8B, 15.48) (Qwen2.5-7B, 14.69)};
    % 手动添加数值标签
    \node[font=\tiny, yshift=3pt, xshift=-1pt] at (axis cs:LLaMA-3.1-8B, 15.48) {15.48};
    \node[font=\tiny, yshift=-3pt, xshift=-1pt] at (axis cs:Qwen2.5-7B, 14.69) {14.69};
    
    \addplot[fill=my-yellow, draw=none] coordinates {(LLaMA-3.1-8B, 15.88) (Qwen2.5-7B, 14.71)};
    % 手动添加数值标签
    \node[font=\tiny, yshift=3pt, anchor=west] at (axis cs:LLaMA-3.1-8B, 15.88) {15.88};
    \node[font=\tiny, yshift=3pt, anchor=west] at (axis cs:Qwen2.5-7B, 14.71) {14.71};
    
    \end{axis}
    }
    \end{tikzpicture}
    \caption{Performance comparisons of utilizing feed forward activations, self-attention activations and whole decoder block activations for cross-lingual connection on \textit{XNLI} and \textit{MKQA} datasets.
    }
    \label{fig:ffn_adv}
\end{figure}
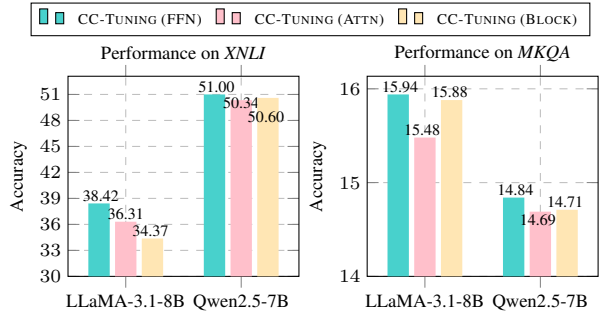

\paragraph{(3) Feed-forward activations contribute the most in cross-lingual connection.}
In addition to investigating cross-lingual connections at the feed-forward activation level, we also explored the potential contributions of self-attention activations and whole decoder block activations. Our results, as shown in Figure~\ref{fig:ffn_adv}, indicate that feed-forward activations have the most pronounced impact on cross-lingual connections within the CC-Tuning paradigm.
This finding highlights the crucial role of feed-forward activations in facilitating cross-lingual latent interactions, which well match the findings presented in \citet{dai-etal-2022-knowledge}, where FFN stores factual knowledge, as well as the motivation of cross-lingual feed forward transplantation operation in \citet{ye2024xtransplant}.
\section{Further Analysis}\label{sec:analysis}

\subsection{Practicality Analysis}

\definecolor{my-red}{RGB}{255,192,203}
\definecolor{my-yellow}{RGB}{255,230,180}
\definecolor{my-blue}{RGB}{0,0,139}
\definecolor{my-green}{RGB}{72,209,204}
\definecolor{my-purple}{RGB}{128,0,128}
\definecolor{my-pink}{RGB}{255,20,147}

\begin{figure}[htbp]
  \centering
  \begin{tikzpicture}[]
      \scriptsize{
      \begin{axis}[
      title={MSE curves},
      title style={yshift=-0.5em},
      at={(12em,0em)}, 
      anchor=north west,  
      ymajorgrids,
      grid style=densely dashed,
      width=0.5\textwidth,
      height=.32\textwidth,
      legend style={
        at={(0.5,1.12)},
        anchor=south,
        legend columns=3,
        nodes={scale=0.8, transform shape},
        column sep=1em,
        },
      xlabel={\scriptsize{Num of \textit{M} Set}},
      ylabel={\scriptsize{MSE value}},
      ylabel style={yshift=0.25em},xlabel style={yshift=0.05em},
      yticklabel style={
            /pgf/number format/fixed, % 强制使用常规小数
            /pgf/number format/precision=2  % 显示两位小数
        },
      ymin=0,ymax=0.12, ytick={0, 0.01, 0.02, 0.03, 0.04, 0.05, 0.06, 0.07, 0.08, 0.09, 0.10, 0.11, 0.12},
      xmin=1,xmax=11, xtick={1, 2, 3, 4, 5, 6, 7, 8, 9, 10, 11}, % 等间隔索引
      xticklabels={100,200,300,500,1000,2000,3000,5000,10000,15000,full},
      ]

      % llama + XNLI
      \addplot[myRed,mark=*,thick,mark options={fill=white,draw=myRed,line width=1.0pt}] coordinates {
        (1,0.081268386)
        (2,0.063585526)
        (3,0.029809737)
        (4,0.018763929)
        (5,0.01242682)
        (6,0.009809057)
        (7,0.008777099)
        (8,0.011546992)
        (9,0.009278043)
        (10,0.007286731)
        (11,0.007206105)
      };
      \addlegendentry{XNLI}

      % llama + XStoryCloze
      \addplot[my-yellow,mark=*,thick,mark options={fill=white,draw=my-yellow,line width=1.0pt}] coordinates {
        (1,0.074566552)
        (2,0.057869516)
        (3,0.030180184)
        (4,0.018980718)
        (5,0.01325627)
        (6,0.010597453)
        (7,0.009516814)
        (8,0.012762878)
        (9,0.010390989)
        (10,0.00829154)
        (11,0.008241931)
      };
      \addlegendentry{XStoryCloze}

      % llama + MMMLU
      \addplot[my-blue,mark=*,thick,mark options={fill=white,draw=my-blue,line width=1.0pt}] coordinates {
        (1,0.094400645)
        (2,0.073906515)
        (3,0.039458129)
        (4,0.0194183)
        (5,0.013195842)
        (6,0.010376166)
        (7,0.009217918)
        (8,0.014495609)
        (9,0.01090696)
        (10,0.007921799)
        (11,0.007786365)
        };
      \addlegendentry{MMMLU}

      % llama + MKQA
      \addplot[my-green,mark=*,thick,mark options={fill=white,draw=my-green,line width=1.0pt}] coordinates {
        (1,0.115208608)
        (2,0.093148507)
        (3,0.050558447)
        (4,0.03040368)
        (5,0.02157206)
        (6,0.016528087)
        (7,0.015078787)
        (8,0.018767151)
        (9,0.015443592)
        (10,0.012908611)
        (11,0.012760016)
              };
      \addlegendentry{MKQA}

      % mistral + XQuAD
      \addplot[my-purple,mark=*,thick,mark options={fill=white,draw=my-purple,line width=1.0pt}] coordinates {
        (1,0.096255113)
        (2,0.072582144)
        (3,0.037873383)
        (4,0.023019559)
        (5,0.015428434)
        (6,0.011082582)
        (7,0.010346412)
        (8,0.013219195)
        (9,0.01047444)
        (10,0.008558465)
        (11,0.008443349)
              };
      \addlegendentry{XQuAD}

      % mistral + XLSum
      \addplot[my-pink,mark=*,thick,mark options={fill=white,draw=my-pink,line width=1.0pt}] coordinates {
        (1,0.080938773)
        (2,0.062887062)
        (3,0.033744622)
        (4,0.021039832)
        (5,0.014313862)
        (6,0.011525604)
        (7,0.010195011)
        (8,0.012643709)
        (9,0.01041441)
        (10,0.005669249)
        (11,0.009322965)
      };
      \addlegendentry{XLSum}

      \end{axis}
     }
    \end{tikzpicture}
  \caption{The curves of mean squared error between feed-forward representations in English and the transformed representations after applying the \textit{Transform Matrix}, as the amount of parallel data used to acquire the \textit{Transform Matrix} increases.}
  % \vspace{-0.5\baselineskip}
  \label{fig:matrix_learn}
\end{figure}
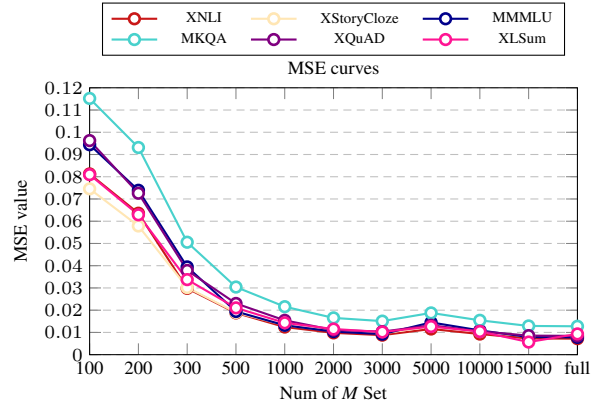

\begin{figure}[t]
\includegraphics[width=1\linewidth]{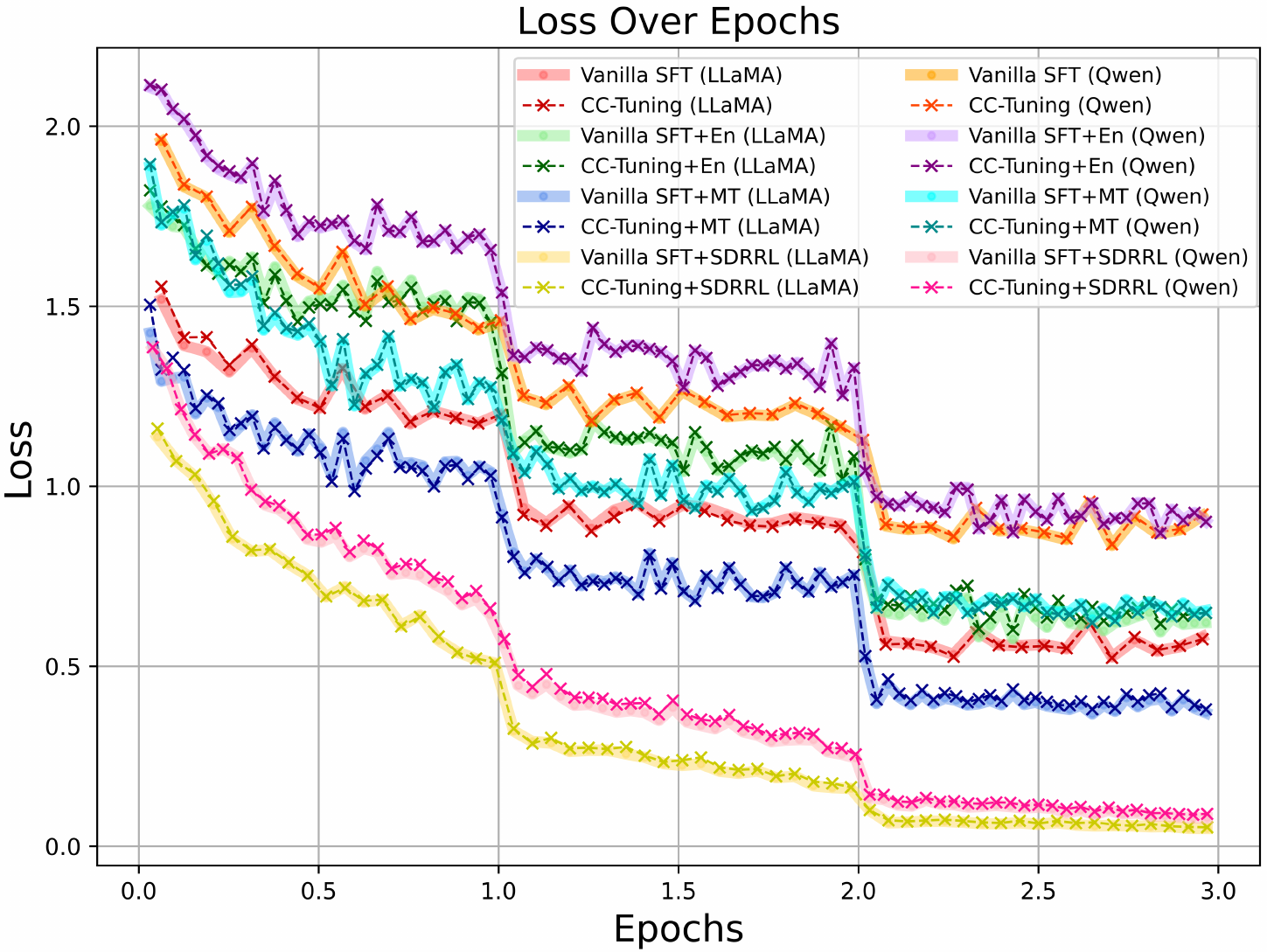}
\caption{The training loss curves of vanilla supervised fine-tuning and \textsc{CC-Tuning} under different training settings (models and training data).}
\label{fig:loss}
\end{figure}

\paragraph{(1) Is the \textit{Transform Matrix} difficult to learn?}
Figure~\ref{fig:matrix_learn} presents the variation in MSE values between $f_{i,l} \cdot W_T$ and $f^{en}_{i,l}$ as the amount of parallel data, $|M|$, used to acquire the \textit{Transform Matrix} increases. We observe that when $|M| = 1000$, the MSE value starts to converge between $0.01$ and $0.02$, and subsequently exhibits a stable trend. This indicates that only a thousand of parallel data are sufficient to effectively align $f_{i,l}$ with $f^{en}_{i,l}$ through the \textit{Transform Matrix}, suggesting that the \textit{Transform Matrix} is relatively easy to learn.

\paragraph{(2) Does incorporating cross-lingual connection substantially interfere with model training and model inference?}
During \textbf{training}, as shown in Figure~\ref{fig:loss}, the loss curves of vanilla SFT and \textsc{CC-Tuning} are closely aligned, suggesting that the incorporation of cross-lingual connection on top of vanilla SFT introduces only negligible interference to the overall training process. This is primarily because no additional training objectives are introduced.
In terms of training overhead, our statistics show that the training time for \textsc{CC-Tuning} is approximately \textbf{\textit{1.12$\sim$1.16 times}} that of vanilla SFT (Table~\ref{tab:train_time}).
Moreover, the additional linear layer \textit{Decision Maker} accounts for only \textbf{\textit{0.0016\%}} and \textbf{\textit{0.0013\%}} of the total parameter count in \textit{LLaMA-3.1-8B} and \textit{Qwen2.5-7B}, respectively—proportions so small that they are practically negligible.
During \textbf{inference}, the time cost for inference with the \textit{Transform Matrix} is also approximately \textbf{\textit{1.1 times}} that of vanilla inference (Table~\ref{tab:infer_time}).

\begin{figure}[t]
\includegraphics[width=1\linewidth]{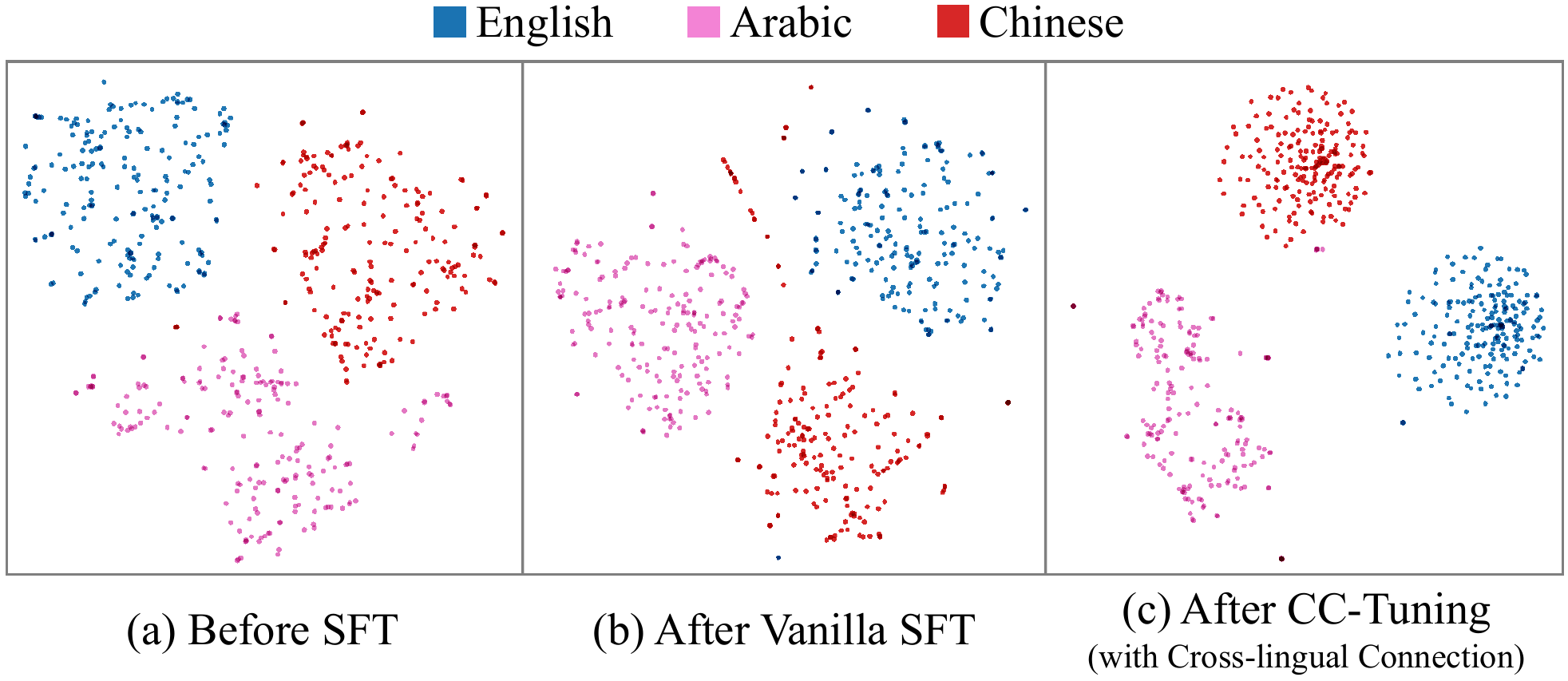}
\caption{t-SNE visualizations of output representations by \textit{LLaMA-3.1-8B} before fine-tuning, after vanilla supervised fine-tuning and after \textsc{CC-Tuning}.}
\label{fig:tSNE}
\vspace{-0.5\baselineskip}
\end{figure}

\subsection{Multilingual Representation Analysis}
To analyze the impact of \textsc{CC-Tuning} on multilingual representations, we employ t-SNE~\citep{van2008visualizing} to visualize the representations of 200 sentences sampled from \textit{XNLI} in parallel across English, Arabic, and Chinese.

As depicted in Figure~\ref{fig:tSNE} (c), after applying \textsc{CC-Tuning}, the multilingual representations show a significantly more compact clustering.
This indicates that \textsc{CC-Tuning} has already facilitated a certain level of cross-lingual interaction through the cross-lingual connection mechanism, allowing the multilingual representations after \textsc{CC-Tuning} require less extensive sharing with representations from other languages in high-dimensional space. And the boundaries between different language representations become more distinct, suggesting that \textsc{CC-Tuning} alleviates the mutual dependency between representations of different languages, enabling the model to exhibit clearer and more distinct multilingual modeling capabilities.

\definecolor{my-blue}{RGB}{100,149,237}  
\definecolor{my-purple}{RGB}{221,160,221}
\definecolor{my-gray}{RGB}{192,192,192}

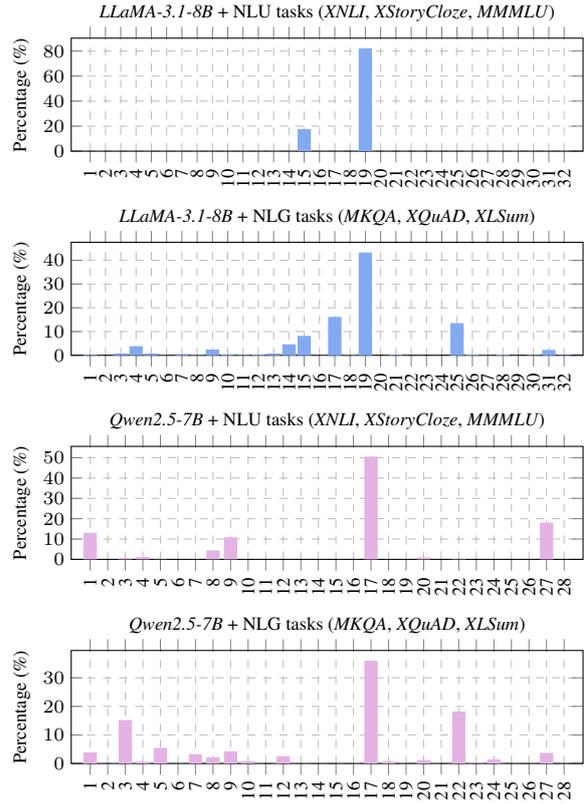
\begin{figure}[t]
\centering
\pgfplotsset{width=1.08\linewidth, height=0.40\linewidth, compat=1.15}
\footnotesize
\begin{tikzpicture}
\scriptsize{
\begin{axis}[
    title={\textit{LLaMA-3.1-8B} + NLU tasks (\textit{XNLI}, \textit{XStoryCloze}, \textit{MMMLU})},
    title style={yshift=-0.3em},
    at={(0em, 0em)},
    ymajorgrids=true,
    xmajorgrids=true,
    grid style=dashed,
    ybar, % 设置柱子宽度
    bar width=5pt, % 设置柱子的宽度
    symbolic x coords={1,2,3,4,5,6,7,8,9,10,11,12,13,14,15,16,17,18,19,20,21,22,23,24,25,26,27,28,29,30,31,32},
    xtick=data,
    xticklabel style={rotate=90, font=\scriptsize}, % 调整 x 轴刻度标签的字体大小
    ymin=0, % Y 轴最小值
    ylabel={Percentage (\%)}, % Y 轴标签
    ytick distance=20, % 设置 Y 轴刻度步长为 2
    ytick align=inside,
    enlarge x limits=0.04, % 增加 X 轴范围，避免柱子被挡住
    ]
    \addplot[fill=my-blue!80, draw=none] coordinates {
        (1, 0.0) (2, 0.0) (3, 0.0) (4, 0.0) (5, 0.0) (6, 0.0) (7, 0.0) (8, 0.0) (9, 0.0) (10, 0.0) (11, 0.0) (12, 0.0) (13, 0.19999999999999998) (14, 0.0) (15, 17.575512905360686) (16, 0.0) (17, 0.13333333333333333) (18, 0.0) (19, 82.09115376130597) (20, 0.0) (21, 0.0) (22, 0.0) (23, 0.0) (24, 0.0) (25, 0.0) (26, 0.0) (27, 0.0) (28, 0.0) (29, 0.0) (30, 0.0) (31, 0.0) (32, 0.0)
    };
    
    \end{axis}

    \begin{axis}[
    title={\textit{LLaMA-3.1-8B} + NLG tasks (\textit{MKQA}, \textit{XQuAD}, \textit{XLSum})},
    title style={yshift=-0.3em},
    at={(0em, -11em)},
    ymajorgrids=true,
    xmajorgrids=true,
    grid style=dashed,
    ybar, % 设置柱子宽度
    bar width=5pt, % 设置柱子的宽度
    symbolic x coords={1,2,3,4,5,6,7,8,9,10,11,12,13,14,15,16,17,18,19,20,21,22,23,24,25,26,27,28,29,30,31,32},
    xtick=data,
    xticklabel style={rotate=90, font=\scriptsize}, % 调整 x 轴刻度标
    ymin=0, % Y 轴最小值
    ylabel={Percentage (\%)}, % Y 轴标签
    ytick distance=10, % 设置 Y 轴刻度步长为 2
    ytick align=inside,
    enlarge x limits=0.04, % 增加 X 轴范围，避免柱子被挡住
    ]
    
    \addplot[fill=my-blue!80, draw=none] coordinates {
        (1, 0.26666666666666666) (2, 0.0) (3, 0.792156862745098) (4, 3.8226890756302523) (5, 0.7333333333333334) (6, 0.061344537815126055) (7, 0.534733893557423) (8, 0.16666666666666666) (9, 2.415406162464986) (10, 0.43333333333333335) (11, 0.19999999999999998) (12, 0.3) (13, 0.7280112044817928) (14, 4.62296918767507) (15, 8.159943977591036) (16, 0.06666666666666667) (17, 16.126890756302522) (18, 0.16666666666666666) (19, 43.26442577030812) (20, 0.0) (21, 0.24537815126050422) (22, 0.03333333333333333) (23, 0.0) (24, 0.0) (25, 13.523249299719888) (26, 0.36666666666666664) (27, 0.09467787114845938) (28, 0.3) (29, 0.0) (30, 0.16666666666666666) (31, 2.274789915966386) (32, 0.13333333333333333)
    };
    
    \end{axis}

    \begin{axis}[
    title={\textit{Qwen2.5-7B} + NLU tasks (\textit{XNLI}, \textit{XStoryCloze}, \textit{MMMLU})},
    title style={yshift=-0.3em},
    at={(0em, -22em)},
    ymajorgrids=true,
    xmajorgrids=true,
    grid style=dashed,
    ybar, % 设置柱子宽度
    bar width=5pt, % 设置柱子的宽度
    symbolic x coords={1,2,3,4,5,6,7,8,9,10,11,12,13,14,15,16,17,18,19,20,21,22,23,24,25,26,27,28},
    xtick=data,
    xticklabel style={rotate=90, font=\scriptsize}, % 调整 x 轴刻度标签的字体大小
    ymin=0, % Y 轴最小值
    ylabel={Percentage (\%)}, % Y 轴标签
    ytick distance=10, % 设置 Y 轴刻度步长为 2
    ytick align=inside,
    enlarge x limits=0.04, % 增加 X 轴范围，避免柱子被挡住
    ]
    \addplot[fill=my-purple!80, draw=none] coordinates {
        (1, 13.047099051400837) (2, 0.0) (3, 0.6107875579086698) (4, 1.1887271122876684) (5, 0.09999999999999999) (6, 0.0) (7, 0.022060445621001543) (8, 4.431016986543129) (9, 10.85373924553276) (10, 0.0) (11, 0.0) (12, 0.0) (13, 0.0) (14, 0.0) (15, 0.0) (16, 0.0) (17, 50.5216412971542) (18, 0.0) (19, 0.0) (20, 0.9755129053606884) (21, 0.0) (22, 0.1553937789543349) (23, 0.0) (24, 0.0) (25, 0.0) (26, 0.0) (27, 18.09402161923671) (28, 0.0)
    };
    
    \end{axis}

    \begin{axis}[
    title={\textit{Qwen2.5-7B} + NLG tasks (\textit{MKQA}, \textit{XQuAD}, \textit{XLSum})},
    title style={yshift=-0.3em},
    at={(0em, -33em)},
    ymajorgrids=true,
    xmajorgrids=true,
    grid style=dashed,
    ybar, % 设置柱子宽度
    bar width=5pt, % 设置柱子的宽度
    symbolic x coords={1,2,3,4,5,6,7,8,9,10,11,12,13,14,15,16,17,18,19,20,21,22,23,24,25,26,27,28},
    xtick=data,
    xticklabel style={rotate=90, font=\scriptsize}, % 调整 x 轴刻度标
    ymin=0, % Y 轴最小值
    ylabel={Percentage (\%)}, % Y 轴标签
    ytick distance=10, % 设置 Y 轴刻度步长为 2
    ytick align=inside,
    enlarge x limits=0.04, % 增加 X 轴范围，避免柱子被挡住
    ]
    
    \addplot[fill=my-purple!80, draw=none] coordinates {
        (1, 3.881232492997199) (2, 0.06666666666666667) (3, 15.215126050420167) (4, 0.6333333333333333) (5, 5.45266106442577) (6, 0.0) (7, 3.256022408963585) (8, 2.1882352941176473) (9, 4.2669467787114845) (10, 0.8333333333333334) (11, 0.0) (12, 2.561344537815126) (13, 0.06666666666666667) (14, 0.03333333333333333) (15, 0.0) (16, 0.03333333333333333) (17, 35.98011204481792) (18, 0.7322128851540616) (19, 0.0) (20, 1.2254901960784315) (21, 0.0) (22, 18.223249299719885) (23, 0.19999999999999998) (24, 1.4282913165266107) (25, 0.0) (26, 0.0) (27, 3.7224089635854347) (28, 0.0)
    };
    \end{axis}
    
    }
    \end{tikzpicture}
    \caption{The distribution of the layer with the highest probability of being selected after the \textit{Decision Maker} over NLU and NLG tasks.
    }
    \label{fig:benefit}
    \vspace{-0.5\baselineskip}
\end{figure}

\subsection{Beneficial Layer Distribution Analysis}
In this section, we present the distribution of the layer with the highest probability of being selected by the \textit{Decision Maker} across NLU and NLG tasks, as shown in Figure~\ref{fig:benefit}. This analysis explores layer-wise effectiveness within the cross-lingual connection. The distribution results indicate that LLMs tend to predominantly utilize the middle layers for both NLU and NLG tasks (\textit{LLaMA-3.1-8B}: 19; \textit{Qwen2.5-7B}: 17), which suggests that the middle layers may capture more valuable and generalized knowledge, potentially acting as a bridge between representations in different languages. 
Additionally, we observe that the beneficial layers identified in NLG tasks are more diverse, likely due to the inherent complexity of generation tasks. In contrast, NLU tasks—primarily focused on selecting from predefined options (e.g., A, B, C, or D)—are less complex, and thus, the layer distribution tend to be more concentrated.

\begin{table*}[t]
\small
\renewcommand{\arraystretch}{1.2} % 设置行间距为默认的 1.5 倍
\setlength{\dashlinedash}{4pt} % 设置虚线段长度
\setlength{\dashlinegap}{2pt}  % 设置虚线段之间的间隔
  \centering
  \resizebox{\linewidth}{!}{
    \begin{tabular}{lccccccccccc}
    \toprule
    XQuAD (Ask in \textit{X}, Answer in English) & en & ar & de & el & hi & ru & th & tr & vi & zh & AVG \\
    \midrule
    \textsc{Ml-SFT} & 72.61 & 15.29 & 30.84 & 21.60 & 12.18 & 15.21 & 18.82 & 25.38 & 33.87 & 13.95 & 25.97\\
    \textsc{CC-Tuning} & 74.62 & 18.82 & 35.71 & 24.54 & 12.27 & 17.14 & 25.29 & 27.23 & 35.21 & 17.90 & 28.87 \\
    \toprule
    XQuAD (Ask in English, Answer in \textit{X}) & en & ar & de & el & hi & ru & th & tr & vi & zh & AVG \\
    \midrule
    \textsc{Ml-SFT} & 72.61 & 20.34 & 40.50 & 18.91 & 21.09 & 20.08 & 17.48 & 29.24 & 31.93 & 27.48 & 29.97 \\
    \textsc{CC-Tuning} & 75.29 & 18.99 & 40.34 & 20.67 & 23.36 & 20.34 & 19.41 & 29.33 & 33.45 & 28.49 & 30.97
\\
    \bottomrule
    \end{tabular}
  }
  \caption{Results on the cross-lingual QA task with \textit{LLaMA-3.1-8B}. The symbol \textit{X} refers to either the input prompt language or the required response language, as specified by the corresponding configuration.}
  \label{tab:cross}
\end{table*}

\subsection{Performance on Cross-lingual Task}
We further conduct additional experiments evaluating the performance of \textsc{CC-Tuning} in cross-lingual scenarios on \textit{XQuAD} under following settings: (1) ``\textit{X}-to-English'': the question is given in language \textit{X}, and the model is explicitly prompted to respond in English. (2) ``English-to-\textit{X}'': the question is given in English, and the model is explicitly prompted to respond in language \textit{X}.

The results in Table~\ref{tab:cross} show that \textsc{CC-Tuning} outperforms vanilla \textsc{Ml-SFT} in both settings, highlighting its effectiveness in cross-lingual scenarios. The advantage is more pronounced in the \textit{X}-to-English'' setting, where the model is given non-English questions. This aligns with the motivation behind \textsc{CC-Tuning}: the models can benefit more when processing non-English inputs by leveraging the latent activations from English. Moreover, the performance gains on English questions under ``English-to-\textit{X}'' setting are relatively smaller, which is also consistent with the observations under \textsc{+En} setting in Table~\ref{tab:main}.

\subsection{Language Confusion Analysis}
Language confusion refers to the cases where the model fails to consistently response in the user’s desired language, or the appropriate language given the context. Here we employ the \textit{lid.176.bin} model from \textit{fastText}, which can identify 176 languages, to evaluate the alignment between model responses and input languages.

The results in Table~\ref{tab:consistency} show that language confusion phenomenon frequently occured in baseline \textsc{SDRRL}. Since \textsc{SDRRL} is designed to facilitate knowledge distillation from resource-rich to low-resource languages, the training data under this setup often contains inconsistencies between input and output languages. While this issue is partially mitigated through code-switching and the integration of external parallel corpora, we observed that it still frequently causes deviations from the prompt language in model responses, making \textsc{SDRRL} less suitable for generation tasks. In contrast, \textsc{CC-Tuning}, along with other baselines, do not exhibit significant language confusions.
\section{Conclusion}
In this paper, we propose \textsc{CC-Tuning}, a novel multilingual fine-tuning paradigm that establishes a cross-lingual connection mechanism at latent level to address the imbalanced multilingual capabilities of current LLMs.
During training, \textsc{CC-Tuning} fuses the feed forward activations from both English and non-English inputs, enabling the model to benefit from both languages.
During inference, we simulate the cross-lingual connection using only monolingual input through representation transformation techniques.
Extensive experiments across six benchmarks covering 22 languages demonstrate that \textsc{CC-Tuning} outperforms vanilla supervised fine-tuning and serves as a strong latent-level alternative to data-level augmentation approaches.
Our results also highlight the importance of rethinking multilingual training paradigms beyond superficial data manipulation, suggesting that deeper architectural interventions may unlock greater potential in LLMs' multilingual capabilities.
\section*{Limitations}
This work exhibits several limitations worth noting.
Firstly, though several ablation experiments are conducted to validate the benefits of our training paradigm, we believe there is much more to explore and investigate in latent cross-lingual interactions. Such interactions should not only be limited to the form discussed in our work.
Secondly, our experiments were conducted on LLaMA-3.1-8B and Qwen2.5-7B. While these models represent important milestones in open-source LLM development, the evaluation across more LLMs would improve the generalizability of our findings across the broader LLM ecosystem.
Thirdly, due to the computational constraints, we did not conduct comparisons between LLMs of different model sizes (particularly larger models), resulting in a lack of insights into the impact of model capacity on performance.
\section*{Acknowledgements}
Xiaocheng Feng is the corresponding author of this work. We thank the anonymous reviewers for their insightful comments. This work was supported by the National Natural Science Foundation of China (NSFC) (grant 62276078, U22B2059), the Key R\&D Program of Heilongjiang via grant 2022ZX01A32,  and the Fundamental Research Funds for the Central Universities (Grant No.HIT.OCEF.2023018). We also thank Huawei Technologies Co., Ltd for supporting part of the computing resources and funding.

\bibliography{custom}

\begin{thebibliography}{52}
\providecommand{\natexlab}[1]{#1}

\bibitem[{Aharoni et~al.(2019)Aharoni, Johnson, and Firat}]{aharoni-etal-2019-massively}
Roee Aharoni, Melvin Johnson, and Orhan Firat. 2019.
\newblock \href {https://doi.org/10.18653/v1/N19-1388} {Massively multilingual neural machine translation}.
\newblock In \emph{Proceedings of the 2019 Conference of the North {A}merican Chapter of the Association for Computational Linguistics: Human Language Technologies, Volume 1 (Long and Short Papers)}, pages 3874--3884, Minneapolis, Minnesota. Association for Computational Linguistics.

\bibitem[{Anil et~al.(2023)Anil, Dai, Firat, Johnson, Lepikhin, Passos, Shakeri, Taropa, Bailey, Chen et~al.}]{anil2023palm}
Rohan Anil, Andrew~M Dai, Orhan Firat, Melvin Johnson, Dmitry Lepikhin, Alexandre Passos, Siamak Shakeri, Emanuel Taropa, Paige Bailey, Zhifeng Chen, et~al. 2023.
\newblock Palm 2 technical report.
\newblock \emph{arXiv preprint arXiv:2305.10403}.

\bibitem[{Artetxe et~al.(2020)Artetxe, Ruder, and Yogatama}]{artetxe-etal-2020-cross}
Mikel Artetxe, Sebastian Ruder, and Dani Yogatama. 2020.
\newblock \href {https://doi.org/10.18653/v1/2020.acl-main.421} {On the cross-lingual transferability of monolingual representations}.
\newblock In \emph{Proceedings of the 58th Annual Meeting of the Association for Computational Linguistics}, pages 4623--4637, Online. Association for Computational Linguistics.

\bibitem[{Chen et~al.(2024)Chen, Ji, Bogoychev, Kutuzov, Haddow, and Heafield}]{chen-etal-2024-monolingual}
Pinzhen Chen, Shaoxiong Ji, Nikolay Bogoychev, Andrey Kutuzov, Barry Haddow, and Kenneth Heafield. 2024.
\newblock \href {https://aclanthology.org/2024.findings-eacl.90/} {Monolingual or multilingual instruction tuning: Which makes a better alpaca}.
\newblock In \emph{Findings of the Association for Computational Linguistics: EACL 2024}, pages 1347--1356, St. Julian{'}s, Malta. Association for Computational Linguistics.

\bibitem[{Chen et~al.(2023)Chen, Jiang, Chen, Wang, Yu, Chen, Zhang, Liang, Zhang, Zhang et~al.}]{chen2023phoenix}
Zhihong Chen, Feng Jiang, Junying Chen, Tiannan Wang, Fei Yu, Guiming Chen, Hongbo Zhang, Juhao Liang, Chen Zhang, Zhiyi Zhang, et~al. 2023.
\newblock Phoenix: Democratizing chatgpt across languages.
\newblock \emph{arXiv preprint arXiv:2304.10453}.

\bibitem[{Conneau et~al.(2020)Conneau, Khandelwal, Goyal, Chaudhary, Wenzek, Guzm{\'a}n, Grave, Ott, Zettlemoyer, and Stoyanov}]{conneau-etal-2020-unsupervised}
Alexis Conneau, Kartikay Khandelwal, Naman Goyal, Vishrav Chaudhary, Guillaume Wenzek, Francisco Guzm{\'a}n, Edouard Grave, Myle Ott, Luke Zettlemoyer, and Veselin Stoyanov. 2020.
\newblock \href {https://doi.org/10.18653/v1/2020.acl-main.747} {Unsupervised cross-lingual representation learning at scale}.
\newblock In \emph{Proceedings of the 58th Annual Meeting of the Association for Computational Linguistics}, pages 8440--8451, Online. Association for Computational Linguistics.

\bibitem[{Conneau et~al.(2018)Conneau, Rinott, Lample, Williams, Bowman, Schwenk, and Stoyanov}]{conneau2018xnli}
Alexis Conneau, Ruty Rinott, Guillaume Lample, Adina Williams, Samuel~R. Bowman, Holger Schwenk, and Veselin Stoyanov. 2018.
\newblock Xnli: Evaluating cross-lingual sentence representations.
\newblock In \emph{Proceedings of the 2018 Conference on Empirical Methods in Natural Language Processing}. Association for Computational Linguistics.

\bibitem[{Cui et~al.(2023)Cui, Yang, and Yao}]{cui2023efficient}
Yiming Cui, Ziqing Yang, and Xin Yao. 2023.
\newblock Efficient and effective text encoding for chinese llama and alpaca.
\newblock \emph{arXiv preprint arXiv:2304.08177}.

\bibitem[{Dai et~al.(2022)Dai, Dong, Hao, Sui, Chang, and Wei}]{dai-etal-2022-knowledge}
Damai Dai, Li~Dong, Yaru Hao, Zhifang Sui, Baobao Chang, and Furu Wei. 2022.
\newblock \href {https://doi.org/10.18653/v1/2022.acl-long.581} {Knowledge neurons in pretrained transformers}.
\newblock In \emph{Proceedings of the 60th Annual Meeting of the Association for Computational Linguistics (Volume 1: Long Papers)}, pages 8493--8502, Dublin, Ireland. Association for Computational Linguistics.

\bibitem[{Dong et~al.(2023)Dong, Li, Dai, Zheng, Wu, Chang, Sun, Xu, and Sui}]{dong2022survey}
Qingxiu Dong, Lei Li, Damai Dai, Ce~Zheng, Zhiyong Wu, Baobao Chang, Xu~Sun, Jingjing Xu, and Zhifang Sui. 2023.
\newblock \href {https://arxiv.org/abs/2301.00234} {A survey for in-context learning}.
\newblock \emph{ArXiv preprint}, abs/2301.00234.

\bibitem[{Dubey et~al.(2024)Dubey, Jauhri, Pandey, Kadian, Al-Dahle, Letman, Mathur, Schelten, Yang, Fan et~al.}]{dubey2024llama}
Abhimanyu Dubey, Abhinav Jauhri, Abhinav Pandey, Abhishek Kadian, Ahmad Al-Dahle, Aiesha Letman, Akhil Mathur, Alan Schelten, Amy Yang, Angela Fan, et~al. 2024.
\newblock The llama 3 herd of models.
\newblock \emph{arXiv preprint arXiv:2407.21783}.

\bibitem[{Garcia et~al.(2024)Garcia, Paiola, Morelli, Candido, J{\'u}nior, Jodas, Afonso, Guilherme, Penteado, and Papa}]{garcia2024introducing}
Gabriel~Lino Garcia, Pedro~Henrique Paiola, Luis~Henrique Morelli, Giovani Candido, Arnaldo~C{\^a}ndido J{\'u}nior, Danilo~Samuel Jodas, Luis Afonso, Ivan~Rizzo Guilherme, Bruno~Elias Penteado, and Jo{\~a}o~Paulo Papa. 2024.
\newblock Introducing bode: A fine-tuned large language model for portuguese prompt-based task.
\newblock \emph{arXiv preprint arXiv:2401.02909}.

\bibitem[{Hasan et~al.(2021)Hasan, Bhattacharjee, Islam, Mubasshir, Li, Kang, Rahman, and Shahriyar}]{hasan-etal-2021-xl}
Tahmid Hasan, Abhik Bhattacharjee, Md.~Saiful Islam, Kazi Mubasshir, Yuan-Fang Li, Yong-Bin Kang, M.~Sohel Rahman, and Rifat Shahriyar. 2021.
\newblock \href {https://aclanthology.org/2021.findings-acl.413} {{XL}-sum: Large-scale multilingual abstractive summarization for 44 languages}.
\newblock In \emph{Findings of the Association for Computational Linguistics: ACL-IJCNLP 2021}, pages 4693--4703, Online. Association for Computational Linguistics.

\bibitem[{Hendrycks et~al.(2020)Hendrycks, Burns, Basart, Zou, Mazeika, Song, and Steinhardt}]{hendrycks2020measuring}
Dan Hendrycks, Collin Burns, Steven Basart, Andy Zou, Mantas Mazeika, Dawn Song, and Jacob Steinhardt. 2020.
\newblock Measuring massive multitask language understanding.
\newblock \emph{arXiv preprint arXiv:2009.03300}.

\bibitem[{HIT-SCIR(2024)}]{Chinese-Mixtral-8x7B}
HIT-SCIR. 2024.
\newblock Chinese-mixtral-8x7b: An open-source mixture-of-experts llm.
\newblock \url{https://github.com/HIT-SCIR/Chinese-Mixtral-8x7B}.

\bibitem[{Huang et~al.(2025)Huang, Yu, Ma, Zhong, Feng, Wang, Chen, Peng, Feng, Qin et~al.}]{huang2025survey}
Lei Huang, Weijiang Yu, Weitao Ma, Weihong Zhong, Zhangyin Feng, Haotian Wang, Qianglong Chen, Weihua Peng, Xiaocheng Feng, Bing Qin, et~al. 2025.
\newblock A survey on hallucination in large language models: Principles, taxonomy, challenges, and open questions.
\newblock \emph{ACM Transactions on Information Systems}, 43(2):1--55.

\bibitem[{Huo et~al.(2025)Huo, Feng, Huang, Fu, Li, Ye, Zhang, Tu, Tang, Lu et~al.}]{huo2025enhancing}
Wenshuai Huo, Xiaocheng Feng, Yichong Huang, Chengpeng Fu, Baohang Li, Yangfan Ye, Zhirui Zhang, Dandan Tu, Duyu Tang, Yunfei Lu, et~al. 2025.
\newblock Enhancing non-english capabilities of english-centric large language models through deep supervision fine-tuning.
\newblock In \emph{Proceedings of the AAAI Conference on Artificial Intelligence}, volume~39, pages 24185--24193.

\bibitem[{Jang et~al.(2016)Jang, Gu, and Poole}]{jang2016categorical}
Eric Jang, Shixiang Gu, and Ben Poole. 2016.
\newblock Categorical reparameterization with gumbel-softmax.
\newblock \emph{arXiv preprint arXiv:1611.01144}.

\bibitem[{Johnson et~al.(2017)Johnson, Schuster, Le, Krikun, Wu, Chen, Thorat, Vi{\'e}gas, Wattenberg, Corrado, Hughes, and Dean}]{johnson-etal-2017-googles}
Melvin Johnson, Mike Schuster, Quoc~V. Le, Maxim Krikun, Yonghui Wu, Zhifeng Chen, Nikhil Thorat, Fernanda Vi{\'e}gas, Martin Wattenberg, Greg Corrado, Macduff Hughes, and Jeffrey Dean. 2017.
\newblock \href {https://doi.org/10.1162/tacl_a_00065} {{G}oogle`s multilingual neural machine translation system: Enabling zero-shot translation}.
\newblock \emph{Transactions of the Association for Computational Linguistics}, 5:339--351.

\bibitem[{Li et~al.(2023)Li, Koto, Wu, Aji, and Baldwin}]{li2023bactrian}
Haonan Li, Fajri Koto, Minghao Wu, Alham~Fikri Aji, and Timothy Baldwin. 2023.
\newblock Bactrian-x: Multilingual replicable instruction-following models with low-rank adaptation.
\newblock \emph{arXiv preprint arXiv:2305.15011}.

\bibitem[{Lin et~al.(2024)Lin, Ji, Tiedemann, Martins, and Sch{\"u}tze}]{lin2024mala}
Peiqin Lin, Shaoxiong Ji, J{\"o}rg Tiedemann, Andr{\'e}~FT Martins, and Hinrich Sch{\"u}tze. 2024.
\newblock Mala-500: Massive language adaptation of large language models.
\newblock \emph{arXiv preprint arXiv:2401.13303}.

\bibitem[{Lin et~al.(2022)Lin, Mihaylov, Artetxe, Wang, Chen, Simig, Ott, Goyal, Bhosale, Du, Pasunuru, Shleifer, Koura, Chaudhary, O{'}Horo, Wang, Zettlemoyer, Kozareva, Diab, Stoyanov, and Li}]{lin-etal-2022-shot}
Xi~Victoria Lin, Todor Mihaylov, Mikel Artetxe, Tianlu Wang, Shuohui Chen, Daniel Simig, Myle Ott, Naman Goyal, Shruti Bhosale, Jingfei Du, Ramakanth Pasunuru, Sam Shleifer, Punit~Singh Koura, Vishrav Chaudhary, Brian O{'}Horo, Jeff Wang, Luke Zettlemoyer, Zornitsa Kozareva, Mona Diab, Veselin Stoyanov, and Xian Li. 2022.
\newblock \href {https://doi.org/10.18653/v1/2022.emnlp-main.616} {Few-shot learning with multilingual generative language models}.
\newblock In \emph{Proceedings of the 2022 Conference on Empirical Methods in Natural Language Processing}, pages 9019--9052, Abu Dhabi, United Arab Emirates. Association for Computational Linguistics.

\bibitem[{Liu et~al.(2023)Liu, Yuan, Fu, Jiang, Hayashi, and Neubig}]{liu2023pre}
Pengfei Liu, Weizhe Yuan, Jinlan Fu, Zhengbao Jiang, Hiroaki Hayashi, and Graham Neubig. 2023.
\newblock Pre-train, prompt, and predict: A systematic survey of prompting methods in natural language processing.
\newblock \emph{ACM Computing Surveys}, 55(9):1--35.

\bibitem[{Longpre et~al.(2021)Longpre, Lu, and Daiber}]{longpre-etal-2021-mkqa}
Shayne Longpre, Yi~Lu, and Joachim Daiber. 2021.
\newblock \href {https://doi.org/10.1162/tacl_a_00433} {{MKQA}: A linguistically diverse benchmark for multilingual open domain question answering}.
\newblock \emph{Transactions of the Association for Computational Linguistics}, 9:1389--1406.

\bibitem[{Muennighoff et~al.(2022)Muennighoff, Wang, Sutawika, Roberts, Biderman, Scao, Bari, Shen, Yong, Schoelkopf et~al.}]{muennighoff2022crosslingual}
Niklas Muennighoff, Thomas Wang, Lintang Sutawika, Adam Roberts, Stella Biderman, Teven~Le Scao, M~Saiful Bari, Sheng Shen, Zheng-Xin Yong, Hailey Schoelkopf, et~al. 2022.
\newblock Crosslingual generalization through multitask finetuning.
\newblock \emph{arXiv preprint arXiv:2211.01786}.

\bibitem[{Ouyang et~al.(2022)Ouyang, Wu, Jiang, Almeida, Wainwright, Mishkin, Zhang, Agarwal, Slama, Ray et~al.}]{ouyang2022training}
Long Ouyang, Jeffrey Wu, Xu~Jiang, Diogo Almeida, Carroll Wainwright, Pamela Mishkin, Chong Zhang, Sandhini Agarwal, Katarina Slama, Alex Ray, et~al. 2022.
\newblock Training language models to follow instructions with human feedback.
\newblock \emph{Advances in neural information processing systems}, 35:27730--27744.

\bibitem[{Pan et~al.(2024)Pan, Huang, Kang, Liu, Lu, and Cheng}]{pan-etal-2024-g}
Xingyuan Pan, Luyang Huang, Liyan Kang, Zhicheng Liu, Yu~Lu, and Shanbo Cheng. 2024.
\newblock \href {https://doi.org/10.18653/v1/2024.acl-long.821} {{G}-{DIG}: Towards gradient-based {DI}verse and hi{G}h-quality instruction data selection for machine translation}.
\newblock In \emph{Proceedings of the 62nd Annual Meeting of the Association for Computational Linguistics (Volume 1: Long Papers)}, pages 15395--15406, Bangkok, Thailand. Association for Computational Linguistics.

\bibitem[{Pires et~al.(2023)Pires, Abonizio, Almeida, and Nogueira}]{pires2023sabia}
Ramon Pires, Hugo Abonizio, Thales~Sales Almeida, and Rodrigo Nogueira. 2023.
\newblock Sabi{\'a}: Portuguese large language models.
\newblock In \emph{Brazilian Conference on Intelligent Systems}, pages 226--240. Springer.

\bibitem[{Qin et~al.(2024)Qin, Chen, Zhou, Chen, Li, Liao, Li, Che, and Yu}]{qin2024multilingual}
Libo Qin, Qiguang Chen, Yuhang Zhou, Zhi Chen, Yinghui Li, Lizi Liao, Min Li, Wanxiang Che, and Philip~S Yu. 2024.
\newblock Multilingual large language model: A survey of resources, taxonomy and frontiers.
\newblock \emph{arXiv preprint arXiv:2404.04925}.

\bibitem[{Ranaldi et~al.(2024)Ranaldi, Pucci, and Freitas}]{ranaldi-etal-2024-empowering}
Leonardo Ranaldi, Giulia Pucci, and Andre Freitas. 2024.
\newblock \href {https://doi.org/10.18653/v1/2024.findings-acl.473} {Empowering cross-lingual abilities of instruction-tuned large language models by translation-following demonstrations}.
\newblock In \emph{Findings of the Association for Computational Linguistics: ACL 2024}, pages 7961--7973, Bangkok, Thailand. Association for Computational Linguistics.

\bibitem[{Santilli and Rodolà(2023)}]{santilli2023camoscio}
Andrea Santilli and Emanuele Rodolà. 2023.
\newblock \href {https://arxiv.org/abs/2307.16456} {Camoscio: an italian instruction-tuned llama}.
\newblock \emph{Preprint}, arXiv:2307.16456.

\bibitem[{Scao et~al.(2022)Scao, Fan, Akiki, Pavlick, Ili{\'c}, Hesslow, Castagn{\'e}, Luccioni, Yvon, Gall{\'e} et~al.}]{scao2022bloom}
Teven~Le Scao, Angela Fan, Christopher Akiki, Ellie Pavlick, Suzana Ili{\'c}, Daniel Hesslow, Roman Castagn{\'e}, Alexandra~Sasha Luccioni, Fran{\c{c}}ois Yvon, Matthias Gall{\'e}, et~al. 2022.
\newblock Bloom: A 176b-parameter open-access multilingual language model.
\newblock \emph{arXiv preprint arXiv:2211.05100}.

\bibitem[{Shaham et~al.(2024)Shaham, Herzig, Aharoni, Szpektor, Tsarfaty, and Eyal}]{shaham-etal-2024-multilingual}
Uri Shaham, Jonathan Herzig, Roee Aharoni, Idan Szpektor, Reut Tsarfaty, and Matan Eyal. 2024.
\newblock \href {https://doi.org/10.18653/v1/2024.findings-acl.136} {Multilingual instruction tuning with just a pinch of multilinguality}.
\newblock In \emph{Findings of the Association for Computational Linguistics: ACL 2024}, pages 2304--2317, Bangkok, Thailand. Association for Computational Linguistics.

\bibitem[{Shanahan(2022)}]{shanahan2022talking}
Murray Shanahan. 2022.
\newblock \href {https://arxiv.org/abs/2212.03551} {Talking about large language models}.
\newblock \emph{ArXiv preprint}, abs/2212.03551.

\bibitem[{Singh et~al.(2024)Singh, Vargus, Dsouza, Karlsson, Mahendiran, Ko, Shandilya, Patel, Mataciunas, OMahony, Zhang, Hettiarachchi, Wilson, Machado, Moura, Krzemiński, Fadaei, Ergün, Okoh, Alaagib, Mudannayake, Alyafeai, Chien, Ruder, Guthikonda, Alghamdi, Gehrmann, Muennighoff, Bartolo, Kreutzer, Üstün, Fadaee, and Hooker}]{singh2024aya}
Shivalika Singh, Freddie Vargus, Daniel Dsouza, Börje~F. Karlsson, Abinaya Mahendiran, Wei-Yin Ko, Herumb Shandilya, Jay Patel, Deividas Mataciunas, Laura OMahony, Mike Zhang, Ramith Hettiarachchi, Joseph Wilson, Marina Machado, Luisa~Souza Moura, Dominik Krzemiński, Hakimeh Fadaei, Irem Ergün, Ifeoma Okoh, Aisha Alaagib, Oshan Mudannayake, Zaid Alyafeai, Vu~Minh Chien, Sebastian Ruder, Surya Guthikonda, Emad~A. Alghamdi, Sebastian Gehrmann, Niklas Muennighoff, Max Bartolo, Julia Kreutzer, Ahmet Üstün, Marzieh Fadaee, and Sara Hooker. 2024.
\newblock \href {https://arxiv.org/abs/2402.06619} {Aya dataset: An open-access collection for multilingual instruction tuning}.
\newblock \emph{Preprint}, arXiv:2402.06619.

\bibitem[{Tang et~al.(2020)Tang, Tran, Li, Chen, Goyal, Chaudhary, Gu, and Fan}]{tang2020multilingual}
Yuqing Tang, Chau Tran, Xian Li, Peng-Jen Chen, Naman Goyal, Vishrav Chaudhary, Jiatao Gu, and Angela Fan. 2020.
\newblock Multilingual translation with extensible multilingual pretraining and finetuning.
\newblock \emph{arXiv preprint arXiv:2008.00401}.

\bibitem[{Touvron et~al.(2023)Touvron, Lavril, Izacard, Martinet, Lachaux, Lacroix, Rozi{\`e}re, Goyal, Hambro, Azhar et~al.}]{touvron2023llama}
Hugo Touvron, Thibaut Lavril, Gautier Izacard, Xavier Martinet, Marie-Anne Lachaux, Timoth{\'e}e Lacroix, Baptiste Rozi{\`e}re, Naman Goyal, Eric Hambro, Faisal Azhar, et~al. 2023.
\newblock Llama: Open and efficient foundation language models.
\newblock \emph{arXiv preprint arXiv:2302.13971}.

\bibitem[{{\"U}st{\"u}n et~al.(2024){\"U}st{\"u}n, Aryabumi, Yong, Ko, D'souza, Onilude, Bhandari, Singh, Ooi, Kayid et~al.}]{ustun2024aya}
Ahmet {\"U}st{\"u}n, Viraat Aryabumi, Zheng-Xin Yong, Wei-Yin Ko, Daniel D'souza, Gbemileke Onilude, Neel Bhandari, Shivalika Singh, Hui-Lee Ooi, Amr Kayid, et~al. 2024.
\newblock Aya model: An instruction finetuned open-access multilingual language model.
\newblock \emph{arXiv preprint arXiv:2402.07827}.

\bibitem[{Van~der Maaten and Hinton(2008)}]{van2008visualizing}
Laurens Van~der Maaten and Geoffrey Hinton. 2008.
\newblock Visualizing data using t-sne.
\newblock \emph{Journal of machine learning research}, 9(11).

\bibitem[{Wang et~al.(2020)Wang, Lipton, and Tsvetkov}]{wang-etal-2020-negative}
Zirui Wang, Zachary~C. Lipton, and Yulia Tsvetkov. 2020.
\newblock \href {https://doi.org/10.18653/v1/2020.emnlp-main.359} {On negative interference in multilingual models: Findings and a meta-learning treatment}.
\newblock In \emph{Proceedings of the 2020 Conference on Empirical Methods in Natural Language Processing (EMNLP)}, pages 4438--4450, Online. Association for Computational Linguistics.

\bibitem[{Wei et~al.(2022{\natexlab{a}})Wei, Tay, Bommasani, Raffel, Zoph, Borgeaud, Yogatama, Bosma, Zhou, Metzler et~al.}]{wei2022emergent}
Jason Wei, Yi~Tay, Rishi Bommasani, Colin Raffel, Barret Zoph, Sebastian Borgeaud, Dani Yogatama, Maarten Bosma, Denny Zhou, Donald Metzler, et~al. 2022{\natexlab{a}}.
\newblock \href {https://arxiv.org/abs/2206.07682} {Emergent abilities of large language models}.
\newblock \emph{ArXiv preprint}, abs/2206.07682.

\bibitem[{Wei et~al.(2022{\natexlab{b}})Wei, Wang, Schuurmans, Bosma, Xia, Chi, Le, Zhou et~al.}]{wei2022chain}
Jason Wei, Xuezhi Wang, Dale Schuurmans, Maarten Bosma, Fei Xia, Ed~Chi, Quoc~V Le, Denny Zhou, et~al. 2022{\natexlab{b}}.
\newblock Chain-of-thought prompting elicits reasoning in large language models.
\newblock \emph{Advances in Neural Information Processing Systems}, 35:24824--24837.

\bibitem[{Wei et~al.(2023)Wei, Wei, Lin, Li, Zhang, Ren, Li, Wan, Cao, Xie et~al.}]{wei2023polylm}
Xiangpeng Wei, Haoran Wei, Huan Lin, Tianhao Li, Pei Zhang, Xingzhang Ren, Mei Li, Yu~Wan, Zhiwei Cao, Binbin Xie, et~al. 2023.
\newblock Polylm: An open source polyglot large language model.
\newblock \emph{arXiv preprint arXiv:2307.06018}.

\bibitem[{Yang et~al.(2024)Yang, Yang, Zhang, Hui, Zheng, Yu, Li, Liu, Huang, Wei, Lin, Yang, Tu, Zhang, Yang, Yang, Zhou, Lin, Dang, Lu, Bao, Yang, Yu, Li, Xue, Zhang, Zhu, Men, Lin, Li, Xia, Ren, Ren, Fan, Su, Zhang, Wan, Liu, Cui, Zhang, and Qiu}]{qwen2.5}
An~Yang, Baosong Yang, Beichen Zhang, Binyuan Hui, Bo~Zheng, Bowen Yu, Chengyuan Li, Dayiheng Liu, Fei Huang, Haoran Wei, Huan Lin, Jian Yang, Jianhong Tu, Jianwei Zhang, Jianxin Yang, Jiaxi Yang, Jingren Zhou, Junyang Lin, Kai Dang, Keming Lu, Keqin Bao, Kexin Yang, Le~Yu, Mei Li, Mingfeng Xue, Pei Zhang, Qin Zhu, Rui Men, Runji Lin, Tianhao Li, Tingyu Xia, Xingzhang Ren, Xuancheng Ren, Yang Fan, Yang Su, Yichang Zhang, Yu~Wan, Yuqiong Liu, Zeyu Cui, Zhenru Zhang, and Zihan Qiu. 2024.
\newblock Qwen2.5 technical report.
\newblock \emph{arXiv preprint arXiv:2412.15115}.

\bibitem[{Ye et~al.(2023)Ye, Tao, and Kong}]{ye2023language}
Jiacheng Ye, Xijia Tao, and Lingpeng Kong. 2023.
\newblock Language versatilists vs. specialists: An empirical revisiting on multilingual transfer ability.
\newblock \emph{arXiv preprint arXiv:2306.06688}.

\bibitem[{Ye et~al.(2024{\natexlab{a}})Ye, Feng, Feng, Ma, Qin, Xu, Yang, Liu, and Qin}]{ye-etal-2024-globesumm}
Yangfan Ye, Xiachong Feng, Xiaocheng Feng, Weitao Ma, Libo Qin, Dongliang Xu, Qing Yang, Hongtao Liu, and Bing Qin. 2024{\natexlab{a}}.
\newblock \href {https://doi.org/10.18653/v1/2024.emnlp-main.603} {{G}lobe{S}umm: A challenging benchmark towards unifying multi-lingual, cross-lingual and multi-document news summarization}.
\newblock In \emph{Proceedings of the 2024 Conference on Empirical Methods in Natural Language Processing}, pages 10803--10821, Miami, Florida, USA. Association for Computational Linguistics.

\bibitem[{Ye et~al.(2024{\natexlab{b}})Ye, Feng, Feng, Qin, Huang, Huang, Ma, Zhang, Lu, Yan et~al.}]{ye2024xtransplant}
Yangfan Ye, Xiaocheng Feng, Xiachong Feng, Libo Qin, Yichong Huang, Lei Huang, Weitao Ma, Zhirui Zhang, Yunfei Lu, Xiaohui Yan, et~al. 2024{\natexlab{b}}.
\newblock Xtransplant: A probe into the upper bound performance of multilingual capability and culture adaptability in llms via mutual cross-lingual feed-forward transplantation.
\newblock \emph{arXiv preprint arXiv:2412.12686}.

\bibitem[{Zhang et~al.(2023)Zhang, Li, Hauer, Shi, and Kondrak}]{zhang-etal-2023-dont}
Xiang Zhang, Senyu Li, Bradley Hauer, Ning Shi, and Grzegorz Kondrak. 2023.
\newblock \href {https://doi.org/10.18653/v1/2023.emnlp-main.491} {Don{'}t trust {C}hat{GPT} when your question is not in {E}nglish: A study of multilingual abilities and types of {LLM}s}.
\newblock In \emph{Proceedings of the 2023 Conference on Empirical Methods in Natural Language Processing}, pages 7915--7927, Singapore. Association for Computational Linguistics.

\bibitem[{Zhang et~al.(2024)Zhang, Wang, Liu, Wang, Wang, Li, Sun, and Liu}]{zhang2024enhancing}
Yuanchi Zhang, Yile Wang, Zijun Liu, Shuo Wang, Xiaolong Wang, Peng Li, Maosong Sun, and Yang Liu. 2024.
\newblock Enhancing multilingual capabilities of large language models through self-distillation from resource-rich languages.
\newblock \emph{arXiv preprint arXiv:2402.12204}.

\bibitem[{Zhao et~al.(2023)Zhao, Zhou, Li, Tang, Wang, Hou, Min, Zhang, Zhang, Dong et~al.}]{zhao2023survey}
Wayne~Xin Zhao, Kun Zhou, Junyi Li, Tianyi Tang, Xiaolei Wang, Yupeng Hou, Yingqian Min, Beichen Zhang, Junjie Zhang, Zican Dong, et~al. 2023.
\newblock \href {https://arxiv.org/abs/2303.18223} {A survey of large language models}.
\newblock \emph{ArXiv preprint}, abs/2303.18223.

\bibitem[{Zheng et~al.(2024)Zheng, Zhang, Zhang, Ye, Luo, Feng, and Ma}]{zheng2024llamafactory}
Yaowei Zheng, Richong Zhang, Junhao Zhang, Yanhan Ye, Zheyan Luo, Zhangchi Feng, and Yongqiang Ma. 2024.
\newblock \href {http://arxiv.org/abs/2403.13372} {Llamafactory: Unified efficient fine-tuning of 100+ language models}.
\newblock In \emph{Proceedings of the 62nd Annual Meeting of the Association for Computational Linguistics (Volume 3: System Demonstrations)}, Bangkok, Thailand. Association for Computational Linguistics.

\bibitem[{Zhu et~al.(2023)Zhu, Lv, Dong, Yuan, Xu, Huang, Kong, Chen, and Li}]{zhu2023extrapolating}
Wenhao Zhu, Yunzhe Lv, Qingxiu Dong, Fei Yuan, Jingjing Xu, Shujian Huang, Lingpeng Kong, Jiajun Chen, and Lei Li. 2023.
\newblock Extrapolating large language models to non-english by aligning languages.
\newblock \emph{arXiv preprint arXiv:2308.04948}.

\end{thebibliography}

\appendix
\newpage

\section{Experiment Details}
\subsection{Baselines Settings}\label{app:baselines}
This section introduces the details of different training data settings.
\begin{itemize}[leftmargin=*]
\setlength{\parsep}{0pt}
\setlength{\parskip}{0pt}
\item \textbf{\textsc{+En}} combines the original multilingual dataset $D$ with its translated parallel English dataset $D^{en}$, resulting in a total training dataset size of $N + N = 2N$.

\item \textbf{\textsc{+MT}} constructs additional translation task form data by pairing the original multilingual dataset $D$ with its translated parallel English dataset $D^{en}$ as follows:
\begin{promptbox}

    \{\\
    "instruction": "Translate the following sentence from English to Spanish.\textbackslash n The category corresponds to politics.",\\
    "output": "La categoría corresponde a política. "\\
    \}
\end{promptbox}
$N$ pairs of parallel data from $D$ and $D^{en}$ can be constructed into $N$ additional samples of translation task form data, resulting in a total training dataset size of $N + N = 2N$.

\item \textbf{\textsc{+SDRRL}}~\citep{zhang2024enhancing} is a self-distillation-based method that integrates English instruction tuning data and its multilingual code-switched extensions. Additionally, it incorporates partially translated data and completion data for fine-tuning (LLaMA-3.1-8B: data size$\approx$$1.2N$, Qwen2.5-7B: data size$\approx$$1.6N$).
\end{itemize}

\subsection{Datasets and Evaluations}\label{app:dataNeval}
\subsubsection{Datasets}
The language subsets used in the 6 evaluation datasets involved in our experiments and the data size used for each language subset are as follows:

\begin{promptbox}

    \textbf{Involved Languages (10 languages each dataset)}
    \vspace{0.5\baselineskip} \\
    XNLI: en, ar, el, hi, ru, sw, th, tr, ur, zh \\
    
    XStoryCloze: en, ar, es, eu, hi, id, ru, sw, te, zh \\
    
    MMMLU: en, ar, bn, es, hi, id, ko, pt, sw, yo \\
    
    XQuAD: en, ar, de, el, hi, ru, th, tr, vi, zh \\
    
    MKQA: en, ar, de, ja, ko, pt, ru, tr, vi, zh \\

    XLSum: en, ar, fr, hi, id, ru, sw, tr, ur, vi \\

    \textbf{A total of 22 unique languages are involved}
\end{promptbox}

\begin{promptbox}

\textbf{Sample Size}
    \vspace{0.5\baselineskip} \\
    \text{XNLI: $1000 \times 10 = 10000$} (parallel)\\
    \text{XStoryCloze: $1511 \times 10 = 15110$} (parallel)\\
    \text{MMMLU: $1000 \times 10 = 10000$} (parallel)\\
    \text{MKQA: $1000 \times 10 = 10000$} (parallel)\\
    \text{XQuAD: $1190 \times 10 = 11900$} (parallel)\\
    \text{XLSum: $100 \times 10 = 1000$} (non-parallel)
\end{promptbox}

\subsubsection{Evaluations}
\textit{XNLI}, \textit{XStoryCloze}, and \textit{MMMLU} all belong to the multiple-choice category. For these datasets, a model's response is considered correct only if it contains the correct option and excludes all other options.
For the short QA generative dataset \textit{MKQA} and \textit{XQuAD}, a model's answer is deemed correct if the gold answer appears in the model's response.

\subsection{Model Responses with \textbf{\textsc{+SDRRL}}}\label{app:NR}
\begin{figure}[t]
\includegraphics[width=1\linewidth]{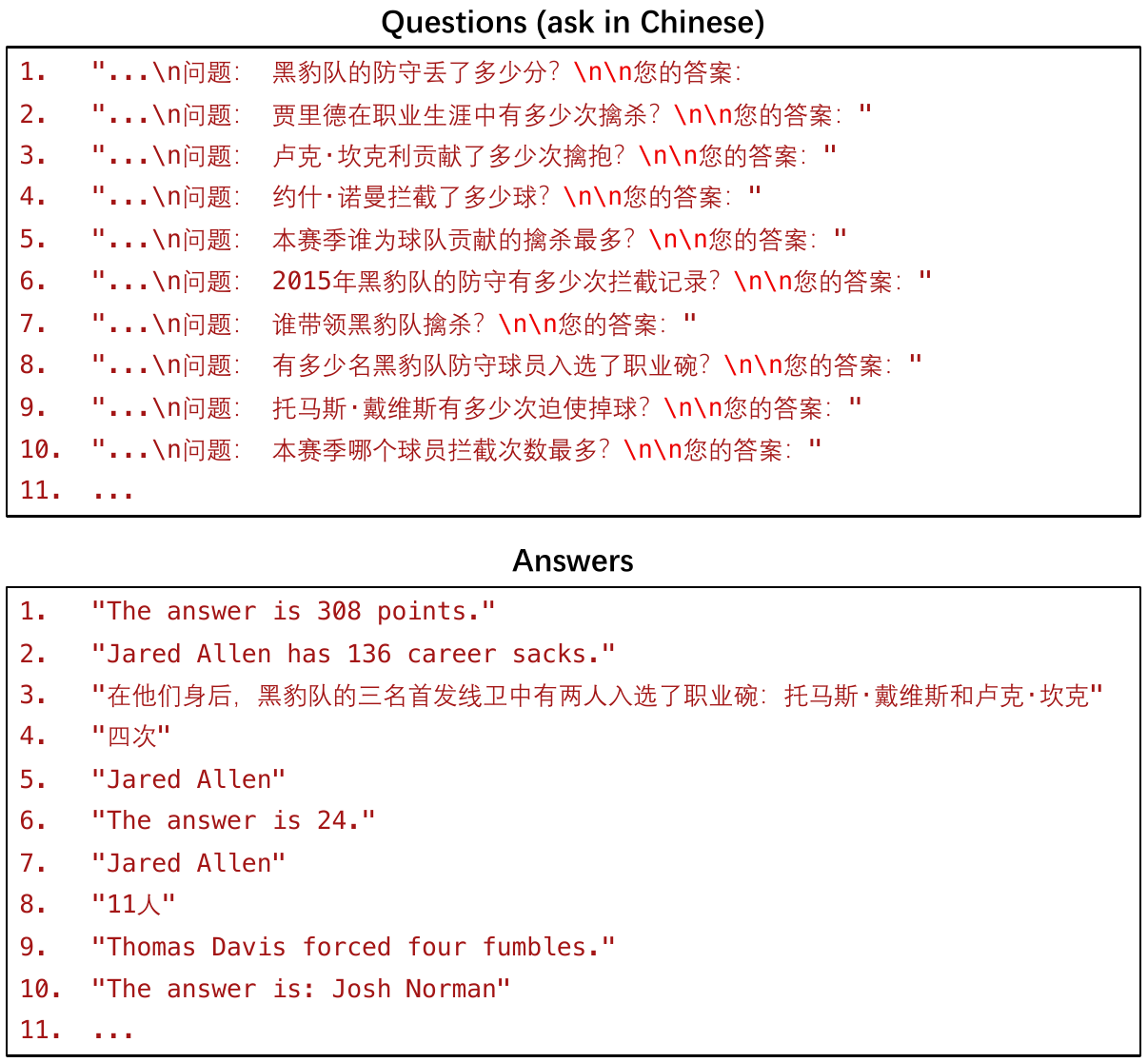}
\caption{Examples of the deviations from the prompt language in model responses when applying \textsc{+SDRRL}.}
\label{fig:SDRRL_cases}
\end{figure}

The results of applying \textbf{\textsc{+SDRRL}} to NLG tasks are not reported in the main body, as it may lead to deviations from the prompt language in model responses. Since \textbf{\textsc{+SDRRL}} aims to achieve distillation from resource-rich languages to low-resource languages, many of the training data's input and output languages under this setup are inconsistent. Although this issue is partially mitigated through code-switching and the incorporation of external parallel corpora, we still observed that it easily leads to deviations from the prompt language in model responses, making it unsuitable for NLG tasks. As in the examples shown in Figure~\ref{fig:SDRRL_cases}, only 3 of the 10 questions given are correctly answered in Chinese, while the rest are all answered in English.

% MSE：Mean Squared Error
\begin{table}[ht]
\small
\renewcommand{\arraystretch}{1.2} % 设置行间距为默认的 1.5 倍
  \centering
  \resizebox{\columnwidth}{!}{
    \begin{tabular}{lcc}
    \toprule
    {Training Time Cost \textit{(h:m:s)}} & \textit{LLaMA-3.1-8B} & \textit{Qwen2.5-7B} \\
    \midrule
    \textsc{Ml-SFT} & 01:36:43 & 01:33:10 \\
    \textsc{CC-Tuning} & 01:51:58 & 01:45:15 \\
    \cdashline{1-3}\noalign{\vskip 0.4ex}
    Time Cost Ratio & 1.16 & 1.13 \\
    \midrule
    \textsc{Ml-SFT+En} & 03:08:25 & 03:03:02 \\
    \textsc{CC-Tuning+En} & 03:34:28 & 03:25:30 \\
    \cdashline{1-3}\noalign{\vskip 0.4ex}
    Time Cost Ratio & 1.14 & 1.12 \\
    \midrule
    \textsc{Ml-SFT+MT} & 03:08:24 & 03:04:13 \\
    \textsc{CC-Tuning+MT} & 03:34:19 & 03:25:59 \\
    \cdashline{1-3}\noalign{\vskip 0.4ex}
    Time Cost Ratio & 1.14 & 1.12 \\
    \midrule
    \textsc{Ml-SFT+SDRRL} & 01:52:17 & 02:23:00 \\
    \textsc{CC-Tuning+SDRRL} & 02:08:52 & 02:41:20 \\
    \cdashline{1-3}\noalign{\vskip 0.4ex}
    Time Cost Ratio & 1.15 & 1.13 \\
    \bottomrule
    \end{tabular}
  }
  \caption{Comparisons of training time cost.}
  \label{tab:train_time}
\end{table}

% MSE：Mean Squared Error
\begin{table}[h]
\small
\renewcommand{\arraystretch}{1.2} % 设置行间距为默认的 1.5 倍
  \centering
  \resizebox{\columnwidth}{!}{
    \begin{tabular}{lcc}
    \toprule
    {Inference Time Cost \textit{(s)}} & \textit{LLaMA-3.1-8B} & \textit{Qwen2.5-7B} \\
    \midrule
    vanilla inference & 2012.26 & 1898.89  \\
    inference w/ \textit{Transform Matrix} & 2209.90 & 2064.50  \\
    \cdashline{1-3}\noalign{\vskip 0.4ex}
    Time Cost Ratio & 1.10 & 1.09 \\
    \bottomrule
    \end{tabular}
  }
  \caption{Comparisons of inference time cost on the Arabic subset of \textit{XNLI} dataset.}
  \label{tab:infer_time}
\end{table}

% MSE：Mean Squared Error
\begin{table}[h]
\small
\renewcommand{\arraystretch}{1.2} % 设置行间距为默认的 1.5 倍
  \centering
  \resizebox{\linewidth}{!}{
    \begin{tabular}{lccc}
    \toprule
    Consistency & MKQA & XQuAD & XLSum \\
    \midrule
    \multicolumn{4}{c}{\textit{LLaMA-3.1-8B}} \\
    \midrule
    Base Model & 0.880 & 0.988 & 0.879 \\
    \textsc{Ml-SFT} & 0.879 & 0.901 & 0.983 \\
    \textsc{Ml-SFT+En} & 0.911 & 0.888 & 0.988 \\
    \textsc{Ml-SFT+MT} & 0.827 & 0.888 & 0.986 \\
    \textsc{Ml-SFT+SDRRL} & 0.360 & 0.447 & 0.352 \\
    \textsc{CC-Tuning} & 0.972 & 0.967 & 0.993 \\
    \midrule
    \multicolumn{4}{c}{\textit{Qwen2.5-7B}} \\
    \midrule
    Base Model & 0.963 & 0.915 & 0.955 \\
    \textsc{Ml-SFT} & 0.996 & 0.999 & 0.999 \\
    \textsc{Ml-SFT+En} & 0.994 & 0.998 & 0.998 \\
    \textsc{Ml-SFT+MT} & 0.995 & 0.998 & 0.981 \\
    \textsc{Ml-SFT+SDRRL} & 0.541 & 0.642 & 0.582 \\
    \textsc{CC-Tuning} & 0.996 & 0.997 & 0.997 \\
    \bottomrule
    \end{tabular}
  }
  \caption{Input and output language consistency results.}
  \label{tab:consistency}
\end{table}

\begin{table*}[ht]
\centering
\scriptsize
\setlength{\tabcolsep}{1pt}
\setlength{\dashlinedash}{3pt} % 设置虚线段长度
\setlength{\dashlinegap}{1pt}  % 设置虚线段之间的间隔
\begin{tabularx}{\textwidth}{p{3.2cm}*{10}{>{\centering\arraybackslash}X}|>{\centering\arraybackslash}X}
    \toprule
    \multicolumn{1}{c}{\bf Models} & \multicolumn{11}{c}{\bf Dataset: XNLI} \\
    \midrule
    & en & ar & el & hi & ru & sw & th & tr & ur & zh & Avg \\
    \midrule
    Vanilla Model (\textit{LLaMA-3.1-8B}) & 36.20 & 15.20 & 23.90 & 31.80 & 29.70 & 28.20 & 29.70 & 28.10 & 24.50 & 8.90 & 25.62 \\
    \midrule
    \textsc{Ml-SFT} (\textit{LLaMA-3.1-8B}) & 12.90 & 35.50 & 35.80 & 31.10 & 34.50 & 31.20 & 31.60 & 37.70 & 33.10 & 35.40 & 31.88 \\
    \cdashline{1-12}\noalign{\vskip 0.4ex}
    \textsc{\quad+En} & 46.60 & 38.70 & 24.40 & 32.20 & 32.70 & 31.90 & 32.00 & 37.60 & 38.90 & 35.20 & 35.02 \\
    \textsc{\quad+MT} & 47.20 & 35.10 & 22.10 & 35.80 & 40.90 & 31.00 & 32.90 & 39.20 & 36.90 & 37.90 & 35.90 \\
    \textsc{\quad+SDRRL} & 29.80 & 29.70 & 29.70 & 29.80 & 29.70 & 29.80 & 29.70 & 29.70 & 29.80 & 29.70 & 29.74 \\
    \midrule
    \textsc{CC-Tuning} (\textit{LLaMA-3.1-8B}) & 51.10 & 40.50 & 38.90 & 33.20 & 42.50 & 30.70 & 37.10 & 39.00 & 35.10 & 36.10 & 38.42 \\
    \cdashline{1-12}\noalign{\vskip 0.4ex}
    \textsc{\quad+En} & 39.50 & 32.00 & 33.40 & 29.20 & 34.50 & 30.00 & 31.20 & 31.70 & 34.50 & 31.20 & 32.72 \\
    \textsc{\quad+MT} & 48.70 & 36.20 & 37.30 & 30.40 & 39.70 & 31.10 & 32.50 & 38.00 & 33.00 & 37.50 & 36.44 \\
    \textsc{\quad+SDRRL} & 29.70 & 29.60 & 29.70 & 31.00 & 29.50 & 29.00 & 29.70 & 29.70 & 30.70 & 29.80 & 29.84 \\
    \midrule
    Vanilla Model (\textit{Qwen2.5-7B}) & 90.20 & 43.60 & 40.40 & 41.80 & 58.70 & 11.60 & 46.90 & 42.50 & 33.60 & 49.00 & 45.83 \\
    \midrule
    \textsc{Ml-SFT} (\textit{Qwen2.5-7B}) & 60.60 & 52.60 & 44.80 & 45.90 & 56.60 & 29.70 & 51.20 & 48.10 & 36.30 & 56.50 & 48.23 \\
    \cdashline{1-12}\noalign{\vskip 0.4ex}
    \textsc{\quad+En} & 81.70 & 54.10 & 39.70 & 43.30 & 59.60 & 30.50 & 50.70 & 49.30 & 39.70 & 59.00 & 50.76 \\
    \textsc{\quad+MT} & 61.60 & 49.60 & 36.70 & 45.80 & 56.80 & 29.70 & 51.60 & 48.10 & 42.00 & 48.60 & 47.05 \\
    \textsc{\quad+SDRRL} & 81.60 & 56.60 & 34.00 & 51.00 & 60.20 & 33.10 & 55.20 & 54.10 & 38.90 & 58.90 & 52.36 \\
    \midrule
    \textsc{CC-Tuning} (\textit{Qwen2.5-7B}) & 78.90 & 51.80 & 41.70 & 44.00 & 60.10 & 30.70 & 52.30 & 50.60 & 40.50 & 59.40 & 51.00 \\
    \cdashline{1-12}\noalign{\vskip 0.4ex}
    \textsc{\quad+En} & 72.00 & 54.30 & 43.10 & 47.00 & 56.50 & 28.60 & 51.70 & 48.40 & 35.90 & 57.30 & 49.48 \\
    \textsc{\quad+MT} & 64.40 & 51.40 & 36.70 & 43.70 & 57.70 & 29.70 & 52.50 & 49.40 & 37.90 & 57.90 & 48.13 \\
    \textsc{\quad+SDRRL} & 83.80 & 56.30 & 33.40 & 46.00 & 60.00 & 32.20 & 58.00 & 58.30 & 42.30 & 60.30 & 53.06 \\
\end{tabularx}

\begin{tabularx}{\textwidth}{p{3.2cm}*{10}{>{\centering\arraybackslash}X}|>{\centering\arraybackslash}X}
    \toprule
    \multicolumn{1}{c}{\bf Models} & \multicolumn{11}{c}{\bf Dataset: XStoryCloze} \\
    \midrule
    & en & ar & es & eu & hi & id & ru & sw & te & zh & Avg \\
    \midrule
    Vanilla Model (\textit{LLaMA-3.1-8B}) & 49.70 & 41.69 & 14.89 & 28.72 & 48.44 & 60.03 & 36.47 & 49.70 & 6.02 & 19.72 & 35.54 \\
    \midrule
    \textsc{Ml-SFT} (\textit{LLaMA-3.1-8B}) & 88.62 & 65.32 & 21.91 & 64.86 & 70.81 & 83.39 & 43.61 & 62.54 & 63.40 & 87.82 & 65.23 \\
    \cdashline{1-12}\noalign{\vskip 0.4ex}
    \textsc{\quad+En} & 77.04 & 40.44 & 65.45 & 59.36 & 77.10 & 79.62 & 49.44 & 60.49 & 55.46 & 86.90 & 65.13 \\
    \textsc{\quad+MT} & 91.73 & 77.70 & 85.04 & 60.82 & 75.78 & 80.48 & 62.14 & 57.84 & 20.91 & 86.57 & 69.90 \\
    \textsc{\quad+SDRRL} & 72.93 & 65.32 & 45.80 & 20.32 & 68.63 & 64.13 & 66.05 & 45.14 & 49.64 & 60.29 & 55.82 \\
    \midrule
    \textsc{CC-Tuning} (\textit{LLaMA-3.1-8B}) & 89.34 & 73.73 & 64.26 & 51.09 & 79.48 & 79.81 & 70.22 & 58.24 & 55.33 & 84.45 & 70.60 \\
    \cdashline{1-12}\noalign{\vskip 0.4ex}
    \textsc{\quad+En} & 69.36 & 66.71 & 75.38 & 25.74 & 62.48 & 75.78 & 49.77 & 50.36 & 49.24 & 84.58 & 60.94 \\
    \textsc{\quad+MT} & 87.43 & 73.99 & 87.62 & 57.91 & 82.06 & 82.86 & 76.44 & 59.43 & 38.65 & 89.01 & 73.54 \\
    \textsc{\quad+SDRRL} & 86.96 & 71.67 & 70.42 & 34.61 & 80.61 & 77.17 & 77.63 & 50.56 & 66.18 & 76.04 & 69.19 \\
    \midrule
    Vanilla Model (\textit{Qwen2.5-7B}) & 85.97 & 85.84 & 91.40 & 18.07 & 78.76 & 69.89 & 91.33 & 17.94 & 55.92 & 75.84 & 67.09 \\
    \midrule
    \textsc{Ml-SFT} (\textit{Qwen2.5-7B}) & 92.12 & 78.89 & 93.51 & 52.95 & 79.48 & 79.48 & 71.21 & 37.06 & 28.92 & 86.96 & 70.06 \\
    \cdashline{1-12}\noalign{\vskip 0.4ex}
    \textsc{\quad+En} & 78.23 & 54.27 & 91.00 & 56.25 & 81.80 & 87.36 & 71.34 & 44.74 & 61.02 & 90.27 & 71.63 \\
    \textsc{\quad+MT} & 82.06 & 56.12 & 92.19 & 57.64 & 82.20 & 88.15 & 73.92 & 29.52 & 57.78 & 85.44 & 70.50 \\
    \textsc{\quad+SDRRL} & 93.85 & 88.42 & 94.51 & 62.61 & 82.00 & 89.15 & 93.05 & 52.88 & 62.01 & 88.22 & 80.67 \\
    \midrule
    \textsc{CC-Tuning} (\textit{Qwen2.5-7B}) & 91.59 & 81.60 & 91.00 & 54.86 & 77.96 & 80.68 & 78.82 & 35.94 & 37.59 & 84.25 & 71.43 \\
    \cdashline{1-12}\noalign{\vskip 0.4ex}
    \textsc{\quad+En} & 39.38 & 37.46 & 90.40 & 55.26 & 79.55 & 86.70 & 85.24 & 45.00 & 42.55 & 85.31 & 64.69 \\
    \textsc{\quad+MT} & 65.32 & 66.64 & 91.66 & 55.06 & 81.67 & 86.43 & 75.12 & 52.75 & 53.47 & 85.77 & 71.39 \\
    \textsc{\quad+SDRRL} & 93.45 & 90.87 & 92.26 & 57.58 & 82.26 & 88.68 & 94.51 & 57.25 & 59.03 & 93.45 & 80.93 \\
\end{tabularx}

\begin{tabularx}{\textwidth}{p{3.2cm}*{10}{>{\centering\arraybackslash}X}|>{\centering\arraybackslash}X}
    \toprule
    \multicolumn{1}{c}{\bf Models} & \multicolumn{11}{c}{\bf Dataset: MMMLU} \\
    \midrule
    & en & ar & bn & es & hi & id & ko & pt & sw & yo & Avg \\
    \midrule
    Vanilla Model (\textit{LLaMA-3.1-8B}) & 45.40 & 28.20 & 17.70 & 11.30 & 25.10 & 26.40 & 25.00 & 11.70 & 16.20 & 0.60 & 20.76 \\
    \midrule
    \textsc{Ml-SFT} (\textit{LLaMA-3.1-8B}) & 57.40 & 41.60 & 31.70 & 51.20 & 37.60 & 44.70 & 39.40 & 51.70 & 20.70 & 26.00 & 40.20 \\
    \cdashline{1-12}\noalign{\vskip 0.4ex}
    \textsc{\quad+En} & 56.80 & 35.90 & 32.00 & 50.90 & 36.00 & 40.80 & 40.50 & 48.10 & 29.80 & 25.40 & 39.62 \\
    \textsc{\quad+MT} & 59.20 & 37.80 & 33.10 & 51.50 & 37.60 & 42.80 & 42.50 & 48.10 & 28.30 & 25.90 & 40.68 \\
    \textsc{\quad+SDRRL} & 53.70 & 32.20 & 23.00 & 27.80 & 35.30 & 30.50 & 27.20 & 36.70 & 12.80 & 1.40 & 28.06 \\
    \midrule
    \textsc{CC-Tuning} (\textit{LLaMA-3.1-8B}) & 57.50 & 41.30 & 33.40 & 51.70 & 37.60 & 43.30 & 41.70 & 46.80 & 27.00 & 27.10 & 40.74 \\
    \cdashline{1-12}\noalign{\vskip 0.4ex}
    \textsc{\quad+En} & 56.50 & 38.10 & 30.80 & 49.90 & 37.00 & 40.70 & 39.00 & 47.30 & 26.80 & 21.20 & 38.73 \\
    \textsc{\quad+MT} & 55.70 & 36.80 & 30.70 & 49.60 & 36.10 & 39.90 & 38.70 & 48.40 & 26.70 & 26.10 & 38.87 \\
    \textsc{\quad+SDRRL} & 53.10 & 38.60 & 33.40 & 46.40 & 36.50 & 41.10 & 35.70 & 47.20 & 31.70 & 14.00 & 37.77 \\
    \midrule
    Vanilla Model (\textit{Qwen2.5-7B}) & 68.20 & 53.50 & 43.70 & 64.00 & 47.40 & 60.90 & 46.00 & 62.40 & 17.90 & 1.30 & 46.53 \\
    \midrule
    \textsc{Ml-SFT} (\textit{Qwen2.5-7B}) & 69.80 & 53.30 & 42.00 & 65.60 & 41.10 & 59.50 & 55.70 & 62.60 & 28.60 & 22.30 & 50.05 \\
    \cdashline{1-12}\noalign{\vskip 0.4ex}
    \textsc{\quad+En} & 65.90 & 53.20 & 37.90 & 64.60 & 40.30 & 56.80 & 55.70 & 62.10 & 28.20 & 23.30 & 48.80 \\
    \textsc{\quad+MT} & 60.50 & 51.40 & 40.00 & 63.90 & 39.50 & 58.30 & 49.00 & 63.00 & 28.80 & 20.50 & 47.49 \\
    \textsc{\quad+SDRRL} & 66.00 & 46.00 & 38.30 & 60.70 & 41.50 & 56.90 & 52.10 & 58.60 & 29.90 & 22.80 & 47.28 \\
    \midrule
    \textsc{CC-Tuning} (\textit{Qwen2.5-7B}) & 69.10 & 52.80 & 40.50 & 65.10 & 41.20 & 59.10 & 54.90 & 62.30 & 30.90 & 20.60 & 49.65 \\
    \cdashline{1-12}\noalign{\vskip 0.4ex}
    \textsc{\quad+En} & 67.10 & 54.50 & 38.10 & 64.20 & 41.90 & 55.50 & 53.90 & 61.30 & 24.10 & 12.90 & 47.35 \\
    \textsc{\quad+MT} & 66.60 & 53.10 & 40.30 & 64.70 & 41.70 & 60.50 & 53.60 & 63.10 & 29.90 & 20.40 & 49.39 \\
    \textsc{\quad+SDRRL} & 66.30 & 50.60 & 39.30 & 60.30 & 41.60 & 56.30 & 52.30 & 57.50 & 28.10 & 26.40 & 47.87 \\
    \bottomrule
\end{tabularx}

\caption{The detailed performance results of different language subsets on NLU tasks (XNLI, XStoryCloze, MMMLU) across all involved models and baselines.}
\label{tab:main_lang_nlu}
\end{table*}

\begin{table*}[ht]
\centering
\scriptsize
\setlength{\tabcolsep}{1pt}
\setlength{\dashlinedash}{3pt} % 设置虚线段长度
\setlength{\dashlinegap}{1pt}  % 设置虚线段之间的间隔
\begin{tabularx}{\textwidth}{p{3.2cm}*{10}{>{\centering\arraybackslash}X}|>{\centering\arraybackslash}X}
    \toprule
    \multicolumn{1}{c}{\bf Models} & \multicolumn{11}{c}{\bf Dataset: MKQA} \\
    \midrule
    & en & ar & de & ja & ko & pt & ru & tr & vi & zh & Avg \\
    \midrule
    Vanilla Model (\textit{LLaMA-3.1-8B}) & 22.50  & 5.20  & 3.50  & 3.50  & 3.00  & 4.80  & 4.90  & 15.70  & 6.80  & 5.70  & 7.56 \\
    \midrule
    \textsc{Ml-SFT} (\textit{LLaMA-3.1-8B}) & 33.60 & 4.90 & 23.30 & 9.10 & 6.00 & 20.70 & 10.00 & 15.60 & 14.30 & 8.90 & 14.64 \\
    \cdashline{1-12}\noalign{\vskip 0.4ex}
    \textsc{\quad+En} & 26.00 & 6.30 & 18.80 & 9.70 & 5.70 & 19.40 & 10.00 & 15.60 & 13.00 & 8.30 & 13.28 \\
    \textsc{\quad+MT} & 29.20 & 5.80 & 19.50 & 9.10 & 5.70 & 17.10 & 10.70 & 15.50 & 14.40 & 8.60 & 13.56 \\
    \textsc{\quad+SDRRL} & -- & -- & -- & -- & -- & -- & -- & -- & -- & -- & -- \\
    \midrule
    \textsc{CC-Tuning} (\textit{LLaMA-3.1-8B}) & 32.00 & 6.00 & 24.10 & 10.90 & 6.30 & 22.40 & 10.50 & 17.80 & 18.20 & 11.20 & 15.94 \\
    \cdashline{1-12}\noalign{\vskip 0.4ex}
    \textsc{\quad+En} & 27.70 & 6.40 & 20.20 & 11.10 & 7.30 & 20.50 & 9.50 & 17.20 & 15.50 & 10.70 & 14.61 \\
    \textsc{\quad+MT} & 32.10 & 6.90 & 21.90 & 10.70 & 7.30 & 21.10 & 10.10 & 17.70 & 17.30 & 10.80 & 15.59 \\
    \textsc{\quad+SDRRL} & -- & -- & -- & -- & -- & -- & -- & -- & -- & -- & -- \\
    \midrule
    Vanilla Model (\textit{Qwen2.5-7B}) & 1.00  & 6.60  & 8.50  & 10.40  & 8.30  & 7.20  & 7.50  & 10.10  & 15.70  & 15.20  & 9.05 \\
    \midrule
    \textsc{Ml-SFT} (\textit{Qwen2.5-7B}) & 30.30 & 6.60 & 19.10 & 11.20 & 8.90 & 19.70 & 9.40 & 12.10 & 15.80 & 14.20 & 14.73 \\
    \cdashline{1-12}\noalign{\vskip 0.4ex}
    \textsc{\quad+En} & 27.80 & 6.90 & 14.80 & 10.80 & 7.40 & 17.50 & 8.10 & 10.10 & 14.70 & 12.40 & 13.05 \\
    \textsc{\quad+MT} & 27.10 & 6.90 & 16.10 & 9.60 & 7.90 & 19.40 & 8.60 & 11.00 & 14.70 & 14.10 & 13.54 \\
    \textsc{\quad+SDRRL} & -- & -- & -- & -- & -- & -- & -- & -- & -- & -- & -- \\
    \midrule
    \textsc{CC-Tuning} (\textit{Qwen2.5-7B}) & 30.3 & 7.2 & 18.60 & 11.6 & 8.5 & 20.5 & 8.3 & 12.9 & 15.80 & 14.7 & 14.84 \\
    \cdashline{1-12}\noalign{\vskip 0.4ex}
    \textsc{\quad+En} & 27.60 & 5.90 & 15.10 & 10.80 & 7.60 & 19.20 & 9.60 & 12.90 & 13.60 & 13.30 & 13.56 \\
    \textsc{\quad+MT} & 29.30 & 6.50 & 16.10 & 11.20 & 7.90 & 18.30 & 8.50 & 12.50 & 13.30 & 14.10 & 13.77 \\
    \textsc{\quad+SDRRL} & -- & -- & -- & -- & -- & -- & -- & -- & -- & -- & -- \\
\end{tabularx}

\begin{tabularx}{\textwidth}{p{3.2cm}*{10}{>{\centering\arraybackslash}X}|>{\centering\arraybackslash}X}
    \toprule
    \multicolumn{1}{c}{\bf Models} & \multicolumn{11}{c}{\bf Dataset: XQuAD} \\
    \midrule
    & en & ar & bn & es & hi & id & ko & pt & sw & yo & Avg \\
    \midrule
    Vanilla Model (\textit{LLaMA-3.1-8B}) & 72.18  & 52.86  & 58.07  & 47.73  & 61.26  & 43.87  & 46.97  & 51.93  & 53.53  & 68.32  & 55.67 \\
    \midrule
    \textsc{Ml-SFT} (\textit{LLaMA-3.1-8B}) & 72.61 & 56.13 & 64.62 & 52.18 & 60.00 & 47.73 & 58.49 & 53.70 & 64.87 & 73.87 & 60.42 \\
    \cdashline{1-12}\noalign{\vskip 0.4ex}
    \textsc{\quad+En} & 63.28 & 53.45 & 62.02 & 51.09 & 57.73 & 46.39 & 58.40 & 50.84 & 61.18 & 69.66 & 57.40 \\
    \textsc{\quad+MT} & 71.76 & 53.78 & 60.84 & 50.84 & 58.99 & 49.33 & 54.03 & 47.73 & 65.21 & 71.51 & 58.40 \\
    \textsc{\quad+SDRRL} & -- & -- & -- & -- & -- & -- & -- & -- & -- & -- & -- \\
    \midrule
    \textsc{CC-Tuning} (\textit{LLaMA-3.1-8B}) & 75.29 & 55.29 & 64.96 & 51.34 & 62.27 & 52.10 & 60.42 & 54.20 & 67.82 & 74.79 & 61.85 \\
    \cdashline{1-12}\noalign{\vskip 0.4ex}
    \textsc{\quad+En} & 69.08 & 58.32 & 64.03 & 52.77 & 60.59 & 51.51 & 60.25 & 52.69 & 66.64 & 73.03 & 60.89 \\
    \textsc{\quad+MT} & 77.73 & 55.29 & 63.45 & 53.61 & 61.34 & 52.18 & 56.72 & 53.11 & 68.24 & 73.78 & 61.55 \\
    \textsc{\quad+SDRRL} & -- & -- & -- & -- & -- & -- & -- & -- & -- & -- & -- \\
    \midrule
    Vanilla Model (\textit{Qwen2.5-7B}) & 53.19  & 71.26  & 71.01  & 50.17  & 49.92  & 56.39  & 64.62  & 57.98  & 77.31  & 89.24  & 64.11 \\
    \midrule
    \textsc{Ml-SFT} (\textit{Qwen2.5-7B}) & 79.92 & 66.97 & 70.08 & 40.00 & 46.39 & 53.95 & 64.96 & 56.05 & 72.77 & 85.04 & 63.61 \\
    \cdashline{1-12}\noalign{\vskip 0.4ex}
    \textsc{\quad+En} & 74.29 & 64.54 & 69.41 & 36.13 & 47.39 & 54.37 & 64.03 & 59.41 & 73.19 & 80.59 & 62.34 \\
    \textsc{\quad+MT} & 79.33 & 65.13 & 69.33 & 41.01 & 50.67 & 52.61 & 67.56 & 58.24 & 72.69 & 83.70 & 64.03 \\
    \textsc{\quad+SDRRL} & -- & -- & -- & -- & -- & -- & -- & -- & -- & -- & -- \\
    \midrule
    \textsc{CC-Tuning} (\textit{Qwen2.5-7B}) & 79.24 & 64.12 & 71.34 & 39.75 & 47.06 & 53.61 & 65.71 & 57.73 & 74.03 & 84.62 & 63.72 \\
    \cdashline{1-12}\noalign{\vskip 0.4ex}
    \textsc{\quad+En} & 72.18 & 64.45 & 68.40 & 41.01 & 47.31 & 54.37 & 66.30 & 58.99 & 72.94 & 80.92 & 62.69 \\
    \textsc{\quad+MT} & 77.98 & 67.31 & 71.26 & 39.75 & 49.41 & 52.86 & 69.33 & 57.82 & 72.27 & 84.62 & 64.26 \\
    \textsc{\quad+SDRRL} & -- & -- & -- & -- & -- & -- & -- & -- & -- & -- & -- \\
\end{tabularx}

\begin{tabularx}{\textwidth}{p{3.2cm}*{10}{>{\centering\arraybackslash}X}|>{\centering\arraybackslash}X}
    \toprule
    \multicolumn{1}{c}{\bf Models} & \multicolumn{11}{c}{\bf Dataset: XLSum} \\
    \midrule
    & en & ar & fr & hi & id & ru & sw & tr & ur & vi & Avg \\
    \midrule
    Vanilla Model (\textit{LLaMA-3.1-8B}) & 6.60  & 3.88  & 11.91  & 1.02  & 5.63  & 7.62  & 3.53  & 5.74  & 1.54  & 9.59  & 5.71 \\
    \midrule
    \textsc{Ml-SFT} (\textit{LLaMA-3.1-8B}) & 24.36 & 9.67 & 18.66 & 1.94 & 13.72 & 14.47 & 8.05 & 11.07 & 6.64 & 14.14 & 12.27 \\
    \cdashline{1-12}\noalign{\vskip 0.4ex}
    \textsc{\quad+En} & 22.46 & 10.62 & 19.66 & 2.97 & 13.72 & 14.02 & 6.76 & 7.14 & 5.77 & 17.27 & 12.04 \\
    \textsc{\quad+MT} & 25.74 & 11.06 & 19.50 & 3.97 & 14.78 & 14.74 & 7.56 & 9.58 & 7.16 & 14.78 & 12.89 \\
    \textsc{\quad+SDRRL} & -- & -- & -- & -- & -- & -- & -- & -- & -- & -- & -- \\
    \midrule
    \textsc{CC-Tuning} (\textit{LLaMA-3.1-8B}) & 25.00 & 10.87 & 19.46 & 3.02 & 13.46 & 15.55 & 8.63 & 10.01 & 7.20 & 15.63 & 12.88 \\
    \cdashline{1-12}\noalign{\vskip 0.4ex}
    \textsc{\quad+En} & 23.76 & 10.26 & 21.45 & 3.67 & 14.30 & 14.45 & 8.94 & 9.94 & 6.15 & 14.92 & 12.78 \\
    \textsc{\quad+MT} & 27.57 & 11.38 & 21.08 & 3.23 & 13.34 & 15.71 & 9.14 & 10.88 & 4.41 & 13.73 & 13.05 \\
    \textsc{\quad+SDRRL} & -- & -- & -- & -- & -- & -- & -- & -- & -- & -- & -- \\
    \midrule
    Vanilla Model (\textit{Qwen2.5-7B}) & 10.45  & 3.59  & 10.86  & 0.00  & 5.43  & 6.89  & 2.73  & 3.54  & 3.09  & 4.21  & 5.08 \\
    \midrule
    \textsc{Ml-SFT} (\textit{Qwen2.5-7B}) & 24.13 & 12.20 & 22.10 & 0.33 & 14.89 & 16.10 & 5.95 & 8.04 & 5.47 & 14.74 & 12.40 \\
    \cdashline{1-12}\noalign{\vskip 0.4ex}
    \textsc{\quad+En} & 23.75 & 11.70 & 20.14 & 0.33 & 14.97 & 15.61 & 6.90 & 8.51 & 5.78 & 14.36 & 12.20 \\
    \textsc{\quad+MT} & 26.72 & 12.32 & 21.47 & 0.67 & 14.00 & 15.29 & 5.66 & 8.73 & 5.12 & 14.78 & 12.48 \\
    \textsc{\quad+SDRRL} & -- & -- & -- & -- & -- & -- & -- & -- & -- & -- & -- \\
    \midrule
    \textsc{CC-Tuning} (\textit{Qwen2.5-7B}) & 23.22 & 10.75 & 22.21 & 0.62 & 14.47 & 17.61 & 6.47 & 8.37 & 5.43 & 15.84 & 12.50 \\
    \cdashline{1-12}\noalign{\vskip 0.4ex}
    \textsc{\quad+En} & 25.06 & 12.79 & 19.58 & 0.33 & 14.71 & 15.63 & 7.12 & 10.62 & 5.39 & 15.01 & 12.63 \\
    \textsc{\quad+MT} & 25.84 & 11.46 & 22.62 & 1.00 & 15.77 & 16.43 & 5.69 & 9.37 & 5.06 & 15.46 & 12.87 \\
    \textsc{\quad+SDRRL} & -- & -- & -- & -- & -- & -- & -- & -- & -- & -- & -- \\
    \bottomrule
\end{tabularx}

\caption{The detailed performance results of different language subsets on NLG tasks (MKQA, XQuAD, XLSum) across all involved models and baselines.}
\label{tab:main_lang_nlg}
\end{table*}

\end{document}